\documentclass{ieeeaccess}
\usepackage{cite}
\usepackage{amsmath,amssymb,amsfonts}
\usepackage{algorithmic}
\usepackage{graphicx}
\usepackage{textcomp}
\usepackage{booktabs}
\usepackage{balance}
\usepackage{lipsum}
\usepackage[utf8]{inputenc}
\usepackage[T1]{fontenc}

\usepackage{bm}
\makeatletter
\AtBeginDocument{\DeclareMathVersion{bold}
\SetSymbolFont{operators}{bold}{T1}{times}{b}{n}
\SetSymbolFont{NewLetters}{bold}{T1}{times}{b}{it}
\SetMathAlphabet{\mathrm}{bold}{T1}{times}{b}{n}
\SetMathAlphabet{\mathit}{bold}{T1}{times}{b}{it}
\SetMathAlphabet{\mathbf}{bold}{T1}{times}{b}{n}
\SetMathAlphabet{\mathtt}{bold}{OT1}{pcr}{b}{n}
\SetSymbolFont{symbols}{bold}{OMS}{cmsy}{b}{n}
\renewcommand\boldmath{\@nomath\boldmath\mathversion{bold}}}
\makeatother

\def\BibTeX{{\rm B\kern-.05em{\sc i\kern-.025em b}\kern-.08em
    T\kern-.1667em\lower.7ex\hbox{E}\kern-.125emX}}

\begin{document}
\history{Date of publication xxxx 00, 0000, date of current version xxxx 00, 0000.}
\doi{XXXX/XXXXXX}

\title{Manipulating Tangible Virtual Object Dynamics to Promote Learning of Precision Force Generation}
\author{\uppercase{Alberto Garz\'as-Villar}\authorrefmark{1,}\authorrefmark{2},
Alba Riera-Cardona\authorrefmark{1}, Alexis Derumigny\authorrefmark{3}, J. Micah Prendergast\authorrefmark{1}, Jane Murray Cramm\authorrefmark{2}, and Laura Marchal-Crespo\authorrefmark{1,}\authorrefmark{4}}

\address[1]{Department of Cognitive Robotics, Delft University of Technology, Delft, 2628 CD, The Netherlands}
\address[2]{Department of Socio-medical Sciences, Erasmus School of Health Policy \& Management, Erasmus University Rotterdam, Rotterdam, 3062PA, The Netherlands.}
\address[3]{Department of Applied Mathematics, Delft University of Technology, Delft, 2628 CD, The Netherlands}
\address[4]{Department of Rehabilitation Medicine, Erasmus Medical Center, Rotterdam, 3015 GD, The Netherlands}
\tfootnote{The authors received financial support from the Convergence Human Mobility Center, a flagship initiative funded by the Convergence Alliance (TU Delft, Erasmus MC, Erasmus University Rotterdam). Grant number: 2022036\\This work was also funded by the Dutch Research Council (NWO, VIDI Grant Nr. 18934)}

\markboth
{Alberto Garz\'as-Villar \headeretal: Manipulating Tangible Virtual Object Dynamics to Promote Learning of Precision Force Generation}
{Alberto Garz\'as-Villar \headeretal: Manipulating Tangible Virtual Object Dynamics to Promote Learning of Precision Force Generation}

\corresp{Corresponding author: Alberto Garz\'as-Villar (e-mail: A.GarzasVillar@tudelft.nl).}

\begin{abstract}
Robotic haptic devices combined with virtual reality offer novel opportunities to train fine force generation, an essential yet overlooked component of post-stroke rehabilitation. This study proposes that manipulating the rendered dynamics of tangible virtual objects can be leveraged to train precise force control while engaging the somatosensory system. 
We conducted an experiment with fifty healthy participants who performed a curling-inspired task in which they had to stretch a virtual spring to generate a target release force to propel the stone to a predefined location on the ice sheet. During training, the spring's force–elongation relationship was modeled as either a linear or non-linear function, i.e., a Gaussian or antisymmetric Gaussian (AS-Gaussian) function with zero derivative at the release target force. Results indicate that the AS-Gaussian group consistently achieved higher force accuracy during training than the linear group, while the Gaussian group only outperformed the linear group toward the end of training. Analysis of personality traits revealed that higher Free Spirit scores were associated with poorer performance and reduced task exploration under Gaussian dynamics, whereas higher Transform-of-Challenge scores correlated with increased exploration. Despite these training effects, no significant differences in long-term retention were found across spring types or personality traits. Participants primarily relied on learned target elongation rather than target force, as evidenced by performance in a transfer task with a different stiffness but the same target force. While promising for somatosensory neurorehabilitation, these methods require refinement to reduce reliance on proprioceptive cues before testing with neurological patients.
\end{abstract}


\titlepgskip=-21pt

\maketitle

\section{Introduction}
\label{Introduction}
\IEEEPARstart{S}{troke} affects around 12 million people yearly worldwide and remains a leading cause of long-term disability~\cite{Feigin2024}. Somatosensory deficits occur in up to 89\% of stroke survivors~\cite{Connell2008}, compromising the integration of proprioceptive and tactile information, which are essential to regulate force generation and control~\cite{Kang2015}. This decline leads to reduced force precision, increased variability, and impacts everyday actions such as grasping, lifting, or manipulating objects~\cite{Quaney2005, Lodha2010}.

Most neurorehabilitation interventions, however, primarily target motor restoration while the somatosensory component, especially somatosensory-guided force control, is generally overlooked~\cite{Bolognini2016,Gassert2018}. The most common approach involves somatosensory stimulation---through electrical~\cite{Lee2024, Mcdonnell2007, Villar2024}, magnetic~\cite{Dafotakis2008,Stefan2008}, or mechanical methods~\cite{Voelcker2010}---to augment sensory processing. Yet, this \textit{passive} stimulation limits patients' active engagement and, therefore, may restrict opportunities for sensorimotor integration and error-driven motor learning~\cite{Krakauer2019}. To overcome this limitation, feedback-based training using visual~\cite{Quaney2010,Kurillo2004} or auditory~\cite{Guo2022} feedback has been proposed to, e.g., support grip force regulation~\cite{Hsu2012}. Yet, reliance on this feedback may reduce the engagement of the somatosensory system, as visual or auditory information alone is often sufficient to perform the task. To our knowledge, the use of somatosensory information to support force control training has mainly targeted the coordination of muscle synergies~\cite{Lum2004, Fischer2006}, whereas its use to train fine force control for arm-object interaction remains under-studied.  \IEEEpubidadjcol

We propose leveraging robots to actively engage patients’ somatosensory systems during force control training. Studies have demonstrated that adaptation to new task dynamics can be accelerated by temporarily modifying those dynamics. For example, the learning rate of walking in a viscous force field could be accelerated when amplifying the force fields during the initial steps, prompting a larger adjustment in muscle commands~\cite{Emken2005}. Studies on force fields in upper-limb tasks~\cite{Patton2004, Patton2006, Rauter2011} further support the idea that temporally modifying task dynamics can provide participants with an enriched sensorimotor experience that challenges the motor system~\cite{Shadmehr1994}. Although promising, previous work focused on manipulating the dynamics of meaningless artificial tasks, e.g., moving in viscous fields, which are far from the force-precision skills needed to perform activities of daily living.

Modulating the rendered dynamics of virtual objects might be a more meaningful way to shape how users generate, regulate, and learn to produce human-object interaction forces. Advanced robotic devices allow precise simulation of interaction forces with tangible virtual objects, known as haptic rendering, and enable programmable, repeatable training environments that cannot be achieved in conventional training~\cite{Ratz2024,Basalp2021}.
Importantly, haptic rendering enables somatosensory engagement in ways that closely mimic natural object manipulation~\cite{Salisbury2004} and support the development of internal models similar to those used in real-world object handling~\cite{Ritter2023}. 

As a proof of concept, we present a system to train fine force generation that manipulates the linearity of the force-elongation relationship of a virtual spring into a Gaussian or antisymmetric Gaussian relationship with zero derivative at the point of correct force generation.  
By doing so, we aim to train---through two different mechanisms---fine force generation leveraging somatosensory feedback without relying on visual or auditory cues. The Gaussian profile introduces a salient deviation from linear dynamics by reducing force beyond the target force, which is expected to promote exploratory behavior and, therefore, movement variability, known to support motor learning \cite{Huang2012, Ozen2021}. 
In contrast, the antisymmetric Gaussian profile provides a localized increase in resistance beyond the target, resembling haptic guidance strategies that constrain movement 
around the desired action, leveraging the exploitation of successful actions as a learning mechanism~\cite{Lum2004, Marchal-Crespo2009}.

We expect that trainees with different personality traits might benefit differently from the proposed modified dynamics. In recent work~\cite{Garzas2025, Garzas2024}, we found that participants with more goal-oriented traits---i.e., autotelic behaviors~\cite{Tse2020} or gaming styles based on achieving goals~\cite{Marczewski2015}---tended to perform worse when training to perform a dynamic task with the support of robotic haptic guidance---i.e., physical guidance through the desired movement---likely because they perceived the guidance as interfering with their task success. Conversely, those participants who exhibited traits more closely related to exploratory behaviors performed worse without guidance, as they probably prioritized exploring the robot behavior over achieving high performance. We also found that participants' Locus of Control---the extent to which individuals attribute outcomes to their own actions versus external factors~\cite{Rotter1966}---influenced how they physically interacted with the robot.

In view of these findings, we designed a study to:  
(1) investigate motor learning of fine force generation under our ``surreal'' haptic rendering-based approach, and (2) determine how individual psychological traits influence users’ interaction with such haptic rendering. Fifty healthy participants with no history of neurological disorders performed a virtual curling-style task using a haptic delta device. Participants were asked to propel a virtual stone across a sliding sheet of ice to reach a defined target position ahead of them. To propel the stone, they had to pull with their arm against a virtual spring rendered by the delta device, and upon release, the stone slid a distance proportional to the force generated by the spring at release. The task is thereby inherently force-based: success depended on selecting and producing the appropriate pulling force at stone release. During training, participants were assigned to one of three groups that experienced different spring dynamics: a standard linear force-elongation relationship, an antisymmetric (AS-) Gaussian, or a Gaussian relationship.

Task performance was quantified as the absolute difference between the target force and force at release and, when using a non-linear spring, the absolute difference between the desired and actual spring elongation at release. We also investigated participants' exploration behavior across the three different virtual springs by comparing the distance the stone traveled before release and the smoothness of its movements. 
Motor learning was assessed as performance improvements from baseline to short- and long-term retention tests under the standard linear dynamics.
To infer whether motor learning relied primarily on force-based information rather than proprioceptive cues, we tested participants' performance in a transfer task that required playing with a linear spring with higher stiffness. 
We also evaluated the influence of personal characteristics on motor performance during training and motor learning. 

We formulated the following hypotheses:

\begin{itemize}
    \item [\textbf{H1}] \textbf{Participant's performance during training will be influenced by the training condition and personality.}
    \begin{itemize}
        \item [\textit{H1.1}] Stiffness profiles that deviate more from the linear spring would increase awareness of the target force and thus promote greater performance gains along training (Gaussian $>$ AS-Gaussian $>$ Linear).
        \item [\textit{H1.2}] Participants with high  goal-oriented characteristics 
        and those reporting higher perceived control over their movements
        , will show greater performance gains when training with the Linear or AS-Gaussian springs compared with average participants, as they would perceive less ``guidance'', in line with~\cite{Garzas2024}.
    \end{itemize}
    \item [\textbf{H2}] \textbf{Exploratory behavior during training will be influenced by the training condition.}
    \begin{itemize}
        \item [\textit{H2.1}] Exploration will decrease over the course of training, irrespective of group assignment.
        \item [\textit{H2.2}] Training conditions that promote larger performance gains in H1.1 will also elicit higher exploration behavior at the beginning of the training.
        \item [\textit{H2.3}] The pronounced deviation of the Gaussian w.r.t. the linear spring is expected to particularly encourage exploration in individuals motivated by novelty and open-ended interaction.
    \end{itemize} 
\end{itemize}

\begin{itemize}
    \item [\textbf{H3}] \textbf{Participants will learn the task differently depending on the training condition and personality.}
    \begin{itemize}
        \item [\textit{H3.1}] Performance will improve from baseline to short- and long-term retention, regardless of the training condition.
        \item [\textit{H3.2}] Performance improvements from baseline to retention will differ across training conditions, reflecting the strategy-dependent differences observed during training (as hypothesized in H1.1).
        \item [\textit{H3.3}] Personality traits associated with improved performance (H1.2) or increased exploration (H2.3) during training will be associated with greater performance gains after training compared to average participants.
    \end{itemize}
    \item [\textbf{H4}] \textbf{During the transfer task}, participants will rely primarily on force perception rather than proprioceptive information. This will lead to a systematic positional offset w.r.t. the ``trained'' target spring elongation.
\end{itemize}

\section{Methods}\label{sec:Methods}
\subsection{Experimental Setup}\label{ExperimentalSetup}
\begin{figure}[h]
    \centering
    \includegraphics[width=\linewidth]{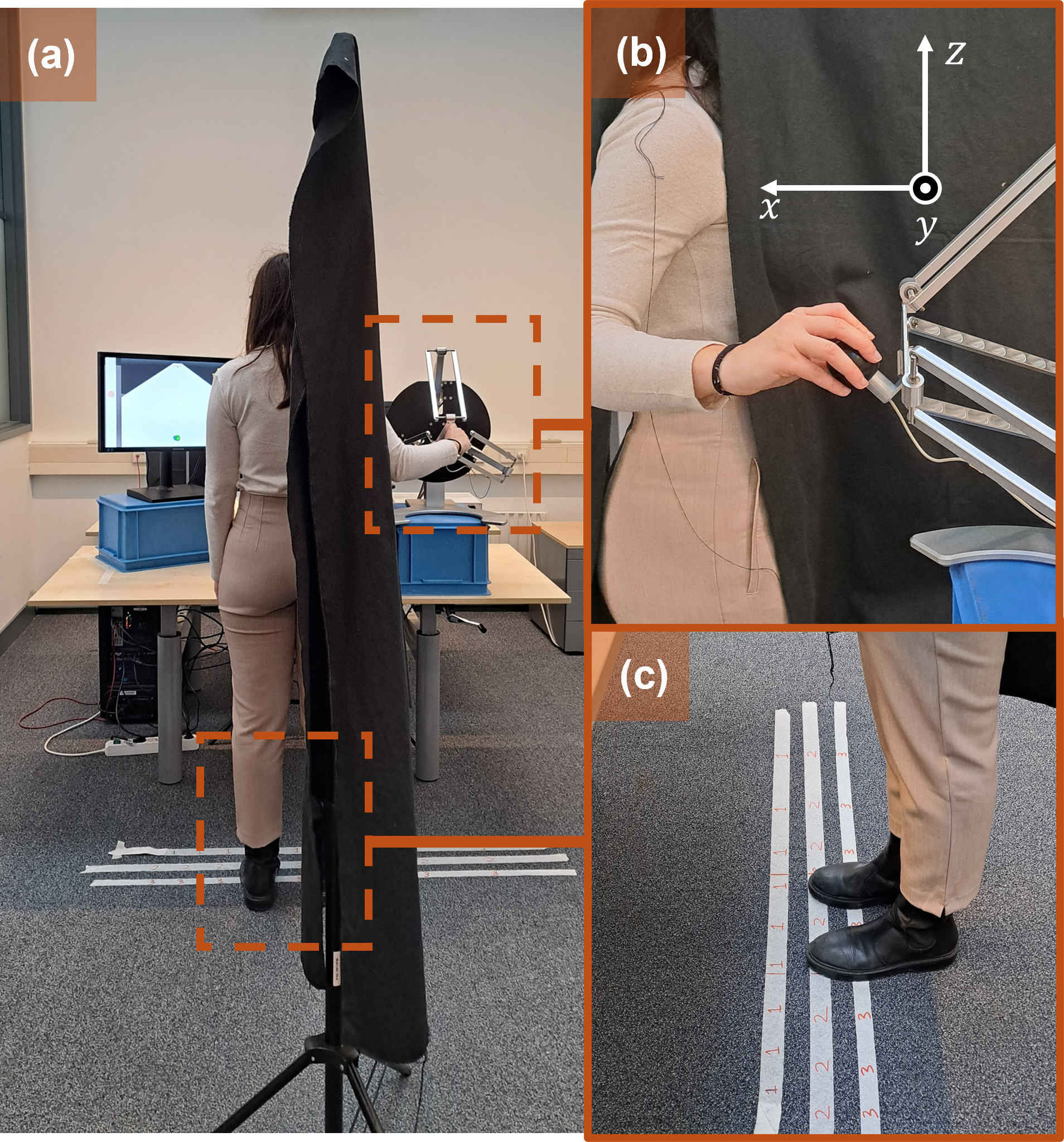}
    \caption{\textbf{(a)} The experimental setup consists of a \textit{Delta.3} haptic device (Force Dimension, Switzerland) together with a computer screen. \textbf{(b)} Holding position of the device end-effector and reference frame. \textbf{(c)} The three predefined feet positions.}
    \label{fig:Setup}
\end{figure}

The experimental setup consisted of a \(Delta.3\) haptic device (Force Dimension, Switzerland) and a computer screen to display the game (Fig.~\ref{fig:Setup}). 
The haptic device provides three translational degrees of freedom at its end-effector and can render forces up to 20\,N along each of the three translational axes. The robot control was implemented in C++. Rendering and data recording were performed at 1\,kHz. Participants interacted with the device via a ball-shaped end-effector equipped with a button.

Both the haptic device and the monitor were positioned on a table, whose height was adjusted so that participants could perform the task while standing in a relaxed posture with the elbow flexed at approximately 90°. For right-handed participants, the monitor was placed on the left and the haptic device on the right, each elevated on dedicated platforms to ensure a comfortable working height. 
For left-handed participants, the setup was mirrored along the $z$-axis (Fig.~\ref{fig:Setup}).

To reduce the influence of visual cues on participants' performance, a curtain was positioned across the participant's shoulder, occluding the device as well as their arm and hand. 
To also reduce the influence of proprioceptive information, we continuously adjusted their position w.r.t. to the robot along the $x$-axis. To do so, we systematically relocate them along three predefined foot-placement references, marked by white stripes on the floor (Fig.~\ref{fig:Setup}c). Each stripe was 4\,cm wide and separated by 3\,cm.

\subsection{The Curling-like Task}\label{TheCurlingGame}
\begin{figure}[h]
    \centering
    \includegraphics[width=\linewidth]{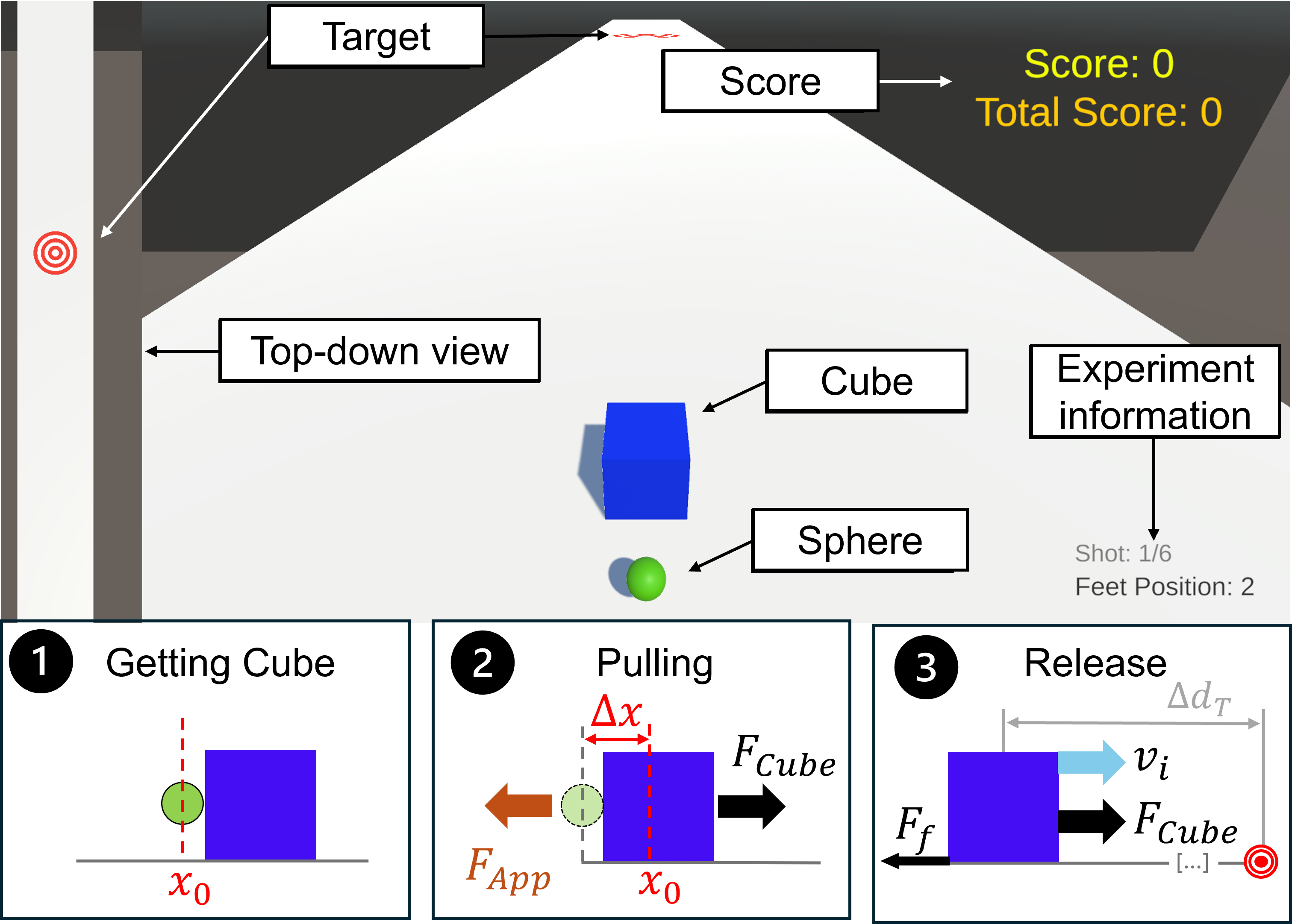}
    \caption{\textit{Up}: Screenshot of the game. \textit{Down}: Representation of the different phases of the cube release. $x_0$: Initial grabbing point, $\Delta x$: Spring elongation, $F_{App}$: Force applied by the participant, $F_{Cube}$: Force generated by the virtual spring, $F_{f}$: Friction force, $v_{i}$: Cube velocity at time $i$, $\Delta d_{T}$: Distance from the cube to the target's center.}
    \label{fig:CurlingGame}
\end{figure}

The force-dependent task was implemented as a virtual curling-style game in Unity3D, version 2022.3.37f1. The virtual environment consisted of a white horizontal surface representing the ice sheet and a blue cube positioned in front of the participant representing the curling stone (Fig.~\ref{fig:CurlingGame}). At the far end of the surface, a target board---red concentric rings---was shown. The goal was to propel the cube using the robot end-effector so that it slid across the surface and stopped as close as possible to the center of the target. On the left side of the screen, a top-down view of the entire surface was displayed to facilitate the visualization of the ``shot'' outcome.

The overall workflow of the task is illustrated in Fig.~\ref{fig:CurlingGame}. Participants moved a green sphere in the game that moved proportionally to the robot end-effector (100\,cm in Unity : 1\,cm robot coordinates). The robot end-effector motion was constrained to translate along the $x$-axis, while two stiff proportional controllers ($k = 7$\,N/m) prevented movements along the other two axes. 
To ``grab'' the virtual cube, participants moved the green sphere until it contacted the cube, then pressed the button on top of the end-effector. Once grabbed, both the sphere and the cube disappeared, eliminating visual cues. 
Participants then pulled the end-effector towards them along the robot's $x$-axis while holding the button pressed. During this movement, the device rendered the force corresponding to the active spring (see Section~\ref{Springs}). The cube reappeared and was propelled as soon as the button was released. At this point, the force experienced by the participant through the device was transmitted to the cube, which in turn slid a distance $\Delta d_T$ on the surface according to the following equation of movement:
\begin{equation}\label{Eq.Fcube}
    F_{Cube} = \frac{m}{UT}\sqrt{2\mu g \Delta d_T},
\end{equation}
where $m=1$\,Kg is the cube’s mass, $g=9.8$\,m/$s^2$ the gravitational acceleration, $\mu=0.02$ the static/dynamic friction coefficient, and UT is the Unity frame update time (0.02\,s). 

The center of the target was located 500\,m from the initial blue block position, with a radius of 29\,m. The scoring system for the six rings, from the innermost to the outermost, was: 100, 20, 10, 5, 2, and 1 points. 
Two scores were displayed on the top-right corner of the screen (Fig.~\ref{fig:CurlingGame}).  
The \textit{Score} reflected the points accumulated within the current trial (see Section~\ref{StudyProtocol}) and was reset at the beginning of each trial, whereas the \textit{Total Score} displayed the cumulative points across the entire experiment. 
Because of the cube's long travel distance after release, its motion was accelerated by a factor of 20. 
Once the cube came to rest due to friction, the points were displayed for 3\,s. The cube then briefly disappeared, only to reappear at the same initial position.

\subsection{Haptic Rendering of the Virtual Springs}\label{Springs}  
\begin{figure}[h]
    \centering
    \includegraphics[width=\linewidth]{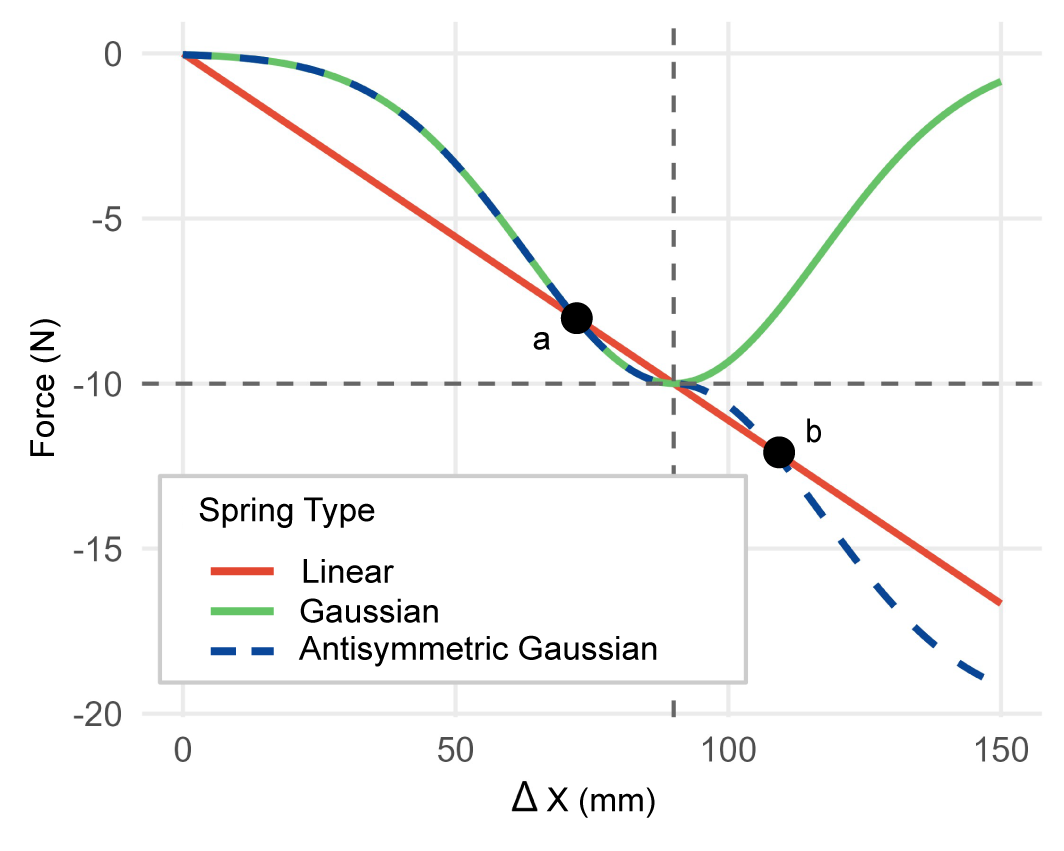}
    \caption{Force-elongation relationship for the different virtual springs. The points \textbf{a} and \textbf{b} indicate the intersections between the Antisymmetric Gaussian and the Linear springs.}
    \label{fig:SpringProfiles}
\end{figure}

The cube was propelled by transferring the force generated by pulling virtual springs. A standard linear spring was used by the \textit{Control} group during training and by all participants in trials not related to training (see Section~\ref{StudyProtocol}). We also defined two non-linear springs to promote learning of the target force that makes the cube reach the center of the target on the floor 
(Fig.~\ref{fig:SpringProfiles}). All virtual springs were designed to reach a target force of $F_T = 10$\,N, corresponding to a target spring elongation $\Delta x_T$ of either 90\,mm (main task), or 70\,mm (transfer task), calculated as the distance between the release point and the initial grabbing point along the robot $x$-axis.

\subsubsection{Linear Spring (LS)}
The LS followed a classical spring equation:
\begin{equation}
    F_{LS} = k_{LS}  \Delta x, \quad \text{where} \quad k_{LS} = \frac{F_T}{\Delta x_T}.
\end{equation}

The force $F_{LS}$ rendered to the participant through the robot end-effector (corresponding to $F_{Cube}$ in Eq.~\eqref{Eq.Fcube}), depends on the cube position w.r.t. the cube's initial grabbing position ($\Delta x$). The stiffness of the LS, $k_{LS}$, is constant and depends on the target force $F_T$ and target elongation $\Delta x_T$ in the corresponding experimental phase.

\subsubsection{Gaussian Spring (GS)}
The GS implemented a Gaussian force–elongation relationship with a zero-derivative at the target force. 
This modification w.r.t. the LS was intended to ``cue'' the target force needed to hit the target center. It was designed to be clearly noticeable and promote active participants' exploration of the spring dynamics.
The force-elongation Gaussian relationship followed the equation:
\begin{equation}
    F_{GS} = F_T \cdot \exp \left(-\frac{(\Delta x - \Delta x_T)^2}{2W^2} \right),
\end{equation}
where $W$ is the width of the Gaussian envelope, set to 27\,mm (in robot end-effector frame).
Note that when the end-effector was pulled beyond the curve’s peak at the target force, the rendered force decreased.

\subsubsection{Anti-symmetric Gaussian Spring (AGS)}
The AGS implemented a nonlinear force–elongation relationship that, in comparison with the GS, introduced a more subtle variation w.r.t. the LS. In particular, the relationship  
evolved following an anti-symmetric Gaussian curve, with the zero-derivative point centered on the target force.

\begin{equation}
\resizebox{0.95\linewidth}{!}{$
F_{AGS} =
\begin{cases}
F_T \cdot \exp\!\left( - \dfrac{(\Delta x - \Delta x_T)^2}{2W^2} \right), & \text{if } \Delta x < \Delta x_T, \\[4pt]
- F_T \cdot \exp\!\left( - \dfrac{(\Delta x - \Delta x_T)^2}{2W^2} \right) + 2F_T, & \text{if } \Delta x \ge \Delta x_T
\end{cases}
$}
\label{eq:FASGS}
\end{equation}

This formulation yielded a force profile with increasing force as the spring elongated, then transitioning to a brief plateau around the target force. Once the plateau was overcome, the force increased steeper than the LS, producing the perceptual effect of a ``subtle wall.''

\subsection{Participants}\label{Participants}
A total of 53 participants completed the experiment, but only 50 could be included in the analysis.
Among the three excluded participants, one did not sign the informed consent, one failed to complete a required questionnaire, and one presented corrupted data.
Of these 50 included participants, 17 identified as female,  33 as male, and none as non-binary. The mean age was 26.3 $\pm$ 3.74\,yo. Handedness was assessed via questionnaire (two left-handed participants). All participants provided written informed consent prior to participation. The study protocol was approved by the TU Delft Human Research Ethics Committee (HREC).

\subsection{Study Protocol}\label{StudyProtocol}
\begin{figure*}[!t]
    \centering
    \includegraphics[width=\linewidth]{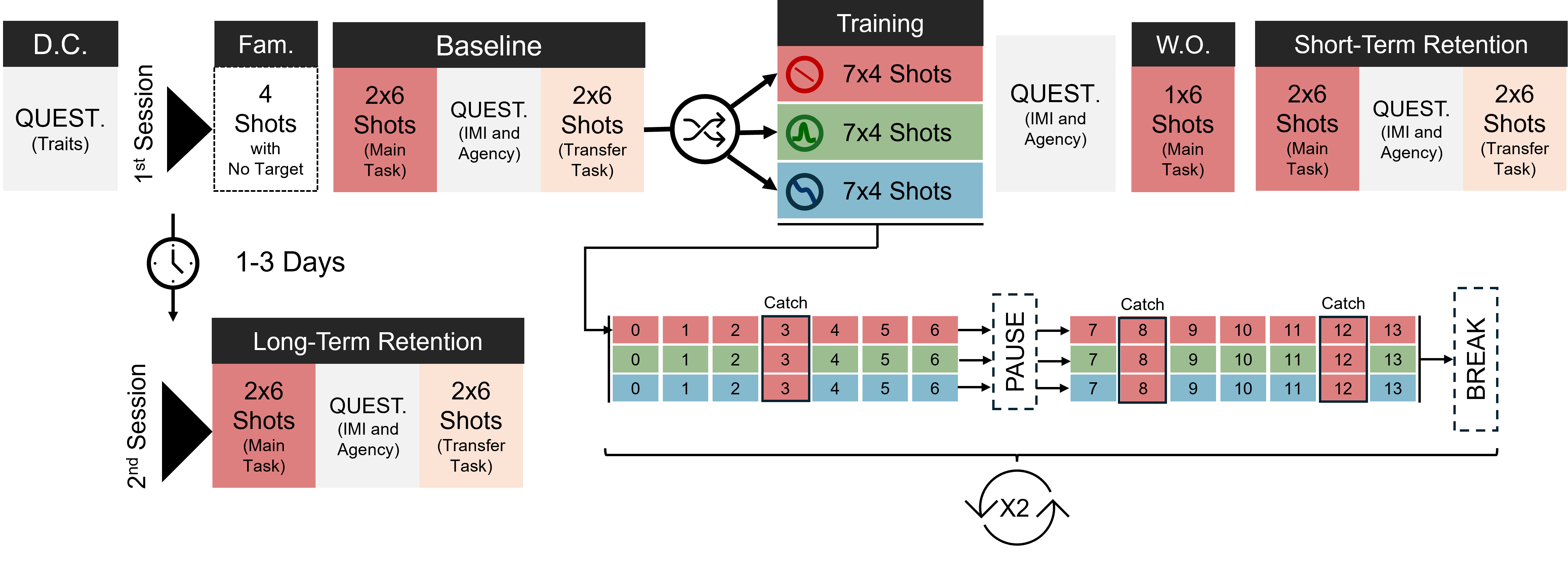}
    \caption{The study protocol included two sessions spaced by 1 to 4 days. A trial comprised 6 shots during the retention phase and 4 during the training phase. D.C.: Data collection, Quest.: Questionnaires, W.O.: Wash out.}
    \label{fig:Protocol}
\end{figure*}

The experiment was conducted at Delft University of Technology, Delft, the Netherlands, and was organized into two sessions, scheduled 1--4 days apart to evaluate long-term retention~\cite{Basalp2021}. Only two participants were scheduled more than four days later, due to personal circumstances.
An overview of the experimental protocol is provided in Fig.~\ref{fig:Protocol}.

Prior to the first session, participants completed an online battery of personality trait questionnaires (Section~\ref{OutcomeMetrics}). 
The remaining data were collected in two separate data collection periods, conducted three months apart, with 30 and 23 participants tested in each period. In both periods, the first half of participants was randomly assigned to each of the three experimental groups, whereas the second half was allocated based on their scores in the Locus of Control Questionnaire (Section~\ref{OutcomeMetrics}) to balance this trait across experimental groups. This pseudo-allocation was implemented based on our previous findings~\cite{Garzas2025}, as Locus of Control could influence how participants react to the potential guiding effects of the different virtual springs, potentially affecting learning. 

The first experimental session (Day 1) began with a five-minute presentation about the curling-like task. Participants were informed that the force felt during cube release would determine its displacement, but were not made aware of the different spring conditions. Participants were also instructed on how to hold the robot end-effector and to press the button using their index finger to reduce variations between participants. 
Participants were then positioned in the experimental setup. The height of the table was adjusted, and they were asked to place their feet 7\,cm from the table, i.e., align with the second white strip on the floor.  
Once the participants were ready, a familiarization phase began. They completed four practice shots with the LS. Each shot consisted of the entire sequence: the cube appearing, grabbing it, releasing it, and waiting until it stopped, allowing participants to understand the game mechanics. No target board was displayed during these familiarization shots. 

After the familiarization phase, the main experiment started with a first Baseline phase to evaluate participants' initial skills. 
During Baseline, they first completed two trials of six shots each with the standard LS (referred to as the \textit{main task}), with no breaks between trials. They then completed a first round of questionnaires, which included items from the Intrinsic Motivation Inventory and an adapted version of the Agency Questionnaire from Piryankova et al.~\cite{Piryankova2014}. Given the focus, amount, and complexity of the present hypotheses, we considered the analysis of these questionnaires out of scope. Participants then performed a Baseline Transfer phase, consisting of two trials of six shots each. In this phase, participants interacted with the transfer spring, an LS with a higher stiffness that shifted the target elongation from 90\,mm to  70\,mm, while the target force was held constant. 
The \textit{transfer task} was incorporated to evaluate whether participants would learn the target force or the target spring elongation.

After Baseline, participants were invited to take a break before starting the Training phase. The Training consisted of four blocks of seven trials each, with each trial comprising four shots. During this phase, the rendered spring was either LS, GS, or AGS, depending on the experimental group each participant was allocated to. After each trial, a message was displayed instructing participants to change their foot placement to a different white strip on the floor to prevent their reliance on proprioception while performing the task (Fig.~\ref{fig:Setup}). The sequence of foot positions (i.e., position 1, 2, and 3) was randomized but kept consistent across participants. To discourage reliance on the training-specific spring, we included a catch trial every 4-5 trials (Fig.~\ref{fig:Protocol}) in which the LS was rendered regardless of group allocation. 
Participants were offered optional breaks between blocks. Only the 2-minute break between the second and third block was mandatory to limit shoulder fatigue. After the last training block, participants were invited to complete the  questionnaires again and rest.

Once participants felt rested, they performed a six-shot Washout trial with the LS to mitigate aftereffects from training with the nonlinear springs. They then proceed to the Short-term Retention phase, which followed the same structure as the Baseline: two trials of six shots each with the LS, followed by a break and the completion of the questionnaires, and two trials of six shots with the transfer spring. The same structure was employed in the Long-term Retention phase performed on the second experimental day.

\subsection{Outcome Metrics}\label{OutcomeMetrics}

\subsubsection{Task Performance}
Participants' performance was assessed by looking at both their accuracy in matching the target force and the consistency with which it was reproduced. Force accuracy was calculated as the absolute value of the \textbf{force error} (in Newtons), defined as the difference between the force at cube release and the target force.
The \textbf{force consistency} (Force SD) was calculated as the standard deviation of the signed force error across the shots within each trial.

The relationship between the spring elongation and the force at release is non-linear in the GS and AGS springs. Therefore, we also evaluated the absolute \textbf{error} between the \textbf{spring elongation} at release and the target spring elongation.

\subsubsection{Exploration Behavior}
To characterize participants’ exploratory behavior during training, we extracted a set of measures from the robot end-effector kinematic data that reflected how individuals interacted with the spring prior to release. Although exploration may take different forms from a behavioral point of view, within this experiment, we expected more exploration to be characterized by longer distance traveled by participants' hands/end-effector, in millimeters, from the onset of cube grasping to its release (\textbf{pre-release path length}). We also expected less smooth movements, calculated as the \textbf{number of directional changes} in the $x$-axis while holding the cube.

\subsubsection{Personality Questionnaires}
Participants completed an online questionnaire administered via Qualtrics (Qualtrix XM, Provo, UT, USA) before the first experimental session. The questionnaire included six personality-related sub-questionnaires. In particular, we included questions related to the \textbf{Transform of Challenge}, \textbf{Transform of Boredom} and \textbf{Curiosity} from the Autotelic Personality Questionnaire~\cite{Tse2020}; the \textbf{Achiever} and \textbf{Free Spirit} elements from the Hexad Gaming Style Questionnaire~\cite{Marczewski2015}; and the \textbf{Locus of Control} (LOC) questionnaire~\cite{Rotter1966}.

All questionnaires, except the LOC, used a seven-point Likert scale. Responses were normalized to values between 0 and 1 based on the theoretical minimum and maximum of each scale, with higher scores indicating stronger expression of the respective trait or gaming style. The LOC questionnaire consisted of 23 multiple-choice items plus six control questions. Raw scores (0 to 23) were linearly transformed to a continuous scale ranging from -1 (\textit{Internal} LOC) to 1 (\textit{External} LOC) to enhance interpretability and facilitate analysis.
Internal LOC reflects the belief that outcomes are contingent on one’s own actions, whereas external LOC attributes outcomes to external forces or chance.

\subsection{Statistical Analysis}\label{StatisticalAnalysis}
The hypotheses (H1--4) outlined in Section~\ref{Introduction} were examined using Linear Mixed-Effects Models (LMMs), implemented with the \texttt{lmer} function from the \texttt{lmerTest} package in \texttt{R}. Statistical significance was defined as $p < 0.05$, and $p$-values were corrected for multiple comparisons using Holm.

\subsubsection{LMM to Analyze Performance During Training (H1)}
We first evaluated the impact of the different spring types and individual personality factors on task performance---namely, absolute force error \& SD and absolute spring elongation error---during training (Hypothesis H1). Since the dataset included many variables and potential interactions, we sequentially compared models using the Akaike Information Criterion (AIC) and the Bayesian Information Criterion (BIC) to reduce the risk of overfitting and select stable models. When competing models showed similar fit, model selection was further guided by effect size comparisons. A complete model comparison is available in Appendix A. The selected model has the form:
\begin{align*}
Var_{per} = \,
&SpringType \times TrialNumber \times{FS}_{c}
 + (1|ID).
\end{align*}

The $SpringType$---i.e., Linear, Gaussian, or AS-Gaussian---and $TrialNumber$---an ordinal from 0 (first training trial) to 27 (last training trial)---were included as independent variables.
Note that the $SpringType$ variable is not equivalent to the assigned training condition, as during catch trials, the Linear spring was rendered to all participants. 

Only one personality trait remained after model reduction: the normalized and centered Free Spirit score (${FS}_{c}$). 
Participant $ID$ was included as a random intercept to account for inter-individual variability.
The dependent variables 
were log-transformed to correct for their skewed distribution due to the absolute values.

\subsubsection{LMM to Analyze Participants' Exploration Behavior During Training (H2)}
To analyze H2, we started with a complete model, as in the previous subsection, with the pre-release path length and number of directional changes as dependent variables. 
The selected model has the form:
\begin{align*}
Var_{beh} = \,
& ({FS}_{c} + {CH}_{c}) \times\\& (SpringType \times  TrialNumber)
 + (1|ID).
\end{align*}

The fixed effects included $SpringType$ and $TrialNumber$, together with the normalized and centered scores from the Transform of Challenge questionnaire ($CH_{c}$) and the Free Spirit questionnaire ($FS_{c}$). 
Participant $ID$ was included as a random intercept to account for inter-individual variability.

\subsubsection{LMM to Analyze Motor Learning (H3)}

To evaluate the impact of the different spring types and individual personality factors on performance improvements across experimental phases, we employed the following model: 
\begin{align*}
Var_{learn} = \,
& ({FS}_{c} + {CU}_{c}) \times\\& (Condition \times Stage + TrialNumber)\\& + Stage \times TrialNumber  
 + (1|ID).
\end{align*}

Here, $Condition$ indicates the group to which participants were allocated (Linear, Gaussian, or AS-Gaussian), and $Stage$ distinguishes between Baseline, STR, and LTR. 
Each $Stage$ was divided into two trials, which was reflected in the variable $TrialNumber$ (0 to 1).
The normalized and centered scores of Free Spirit (${FS}{c}$) and Curiosity (${CU}{c}$) were incorporated as fixed effects after model reduction.
Participant $ID$ was included as a random intercept.

Only the log-transformed absolute force error and the force SD were included as dependent variables since the force–elongation relationship was always linear, and therefore, redundant. Data from the transfer task were not included.

\subsubsection{LMM to Analyze Skill Transfer (H4)}

To evaluate whether participants relied on force-based or positional-based information during the transfer task (H4), we fitted a separate LMM with spring elongation as the dependent variable. We did not use the spring elongation error, as it could lead to a misleading interpretation due to differences in target elongations between the main and transfer springs. 
Because trait- and condition-dependent effects were not of interest for this hypothesis, the resulting model is defined as follows:
\begin{align*}
Var_{Transfer} = \,
& TransTask\times  Stage  \times\\& TrialNumber\times ShotNumber 
 + (1|ID).
\end{align*}

The $TransTask$ indicates whether participants interacted with the main spring (0: target elongation 90\,mm), or the transfer spring (1: target elongation 70\,mm).
The used dataset included data from the Baseline, STR, and LTR phases. Consequently, $TrialNumber$ ranged from 0 to 1. We also included $ShotNumber$ (0 to 5), i.e., the order of shots within a trial, to capture changes across successive shots.

\section{Results}\label{sec:Results}

Complete LMM outputs, including the LMM full tables, are provided in Appendices A and B. To aid interpretation, Figs. \ref{fig:ResultsForceAndPositionTraining} to \ref{fig:ResultTransfer} illustrate predicted changes in the dependent variables and associated 95\% confidence intervals derived from the model estimates. 
Some of the predicted changes include participants with 10\% above or below the average group personality trait, labeled as ``high'' or ``low'' levels, respectively.  
This margin was chosen to facilitate interpretation as it reflects a realistic and meaningful deviation from the average.

\subsection{Task Performance During Training (H1)}\label{Sec:PerformanceDuringTraining}

\subsubsection{Effect of Spring Type in Task Performance (H1.1)}
\begin{figure}[!h]
    \centering
    \includegraphics[width=\linewidth]{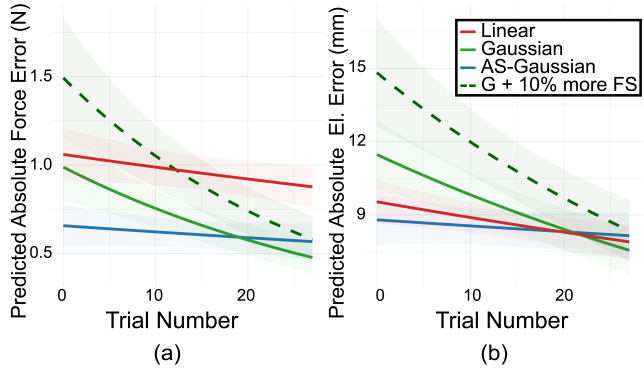}
    \caption{Prediction from the LMM to visualize performance during training. Shaded regions indicate 95\% confidence intervals.
    \textbf{(a)} Evolution of the absolute force error along the 28 training trials. \textbf{(b)} Evolution of the absolute spring elongation error along the 28 training trials.  El.: Spring Elongation, G: Gaussian, FS: Free Spirit}
    \label{fig:ResultsForceAndPositionTraining}
\end{figure}

Regarding \textbf{force accuracy}, participants who trained with the Linear or AS-Gaussian springs did not improve across training trials. However, participants who received the AS-Gaussian spring consistently exhibited lower force error than the linear spring group throughout training while receiving this non-linear spring (Fig.~\ref{fig:ResultsForceAndPositionTraining}a, approximately 38\% reduction; main effect of spring, $p < 0.001 $).  
Participants who trained with the Gaussian spring, on the other hand, showed improvement across the training trials (Fig.~\ref{fig:ResultsForceAndPositionTraining}a, \textit{GaussianSpring × TrialNumber}, $p = 0.011$) and outperformed the linear group by the end of training, with approximately 46\% lower force error in the final trial (pairwise comparison of estimated marginal means at last training trial, $p < 0.001$).

Regarding \textbf{force variability}, a significant reduction over training was observed only for participants trained with the Gaussian spring (\textit{GaussianSpring x TrialNumber}, $p = 0.040$). \textbf{Spring elongation error}, on the other hand, did not show significant effects of spring type or spring–trial interactions, but only a main effect of trial, suggesting a general improvement independent of the training dynamics (Fig.~\ref{fig:ResultsForceAndPositionTraining}b, $p = 0.031$).

\subsubsection{Influence of Personality Traits in Task Performance (H1.2)}

Personality traits modulated task performance during training. Specifically, participants with higher \textbf{Free Spirit} scores showed worse initial performance when training with the Gaussian spring. A participant scoring 10\% higher Free Spirit, for example, was predicted to show 51\% more force error and 29\% more spring elongation error at the beginning of training compared with an average participant training with the same spring (Fig.~\ref{fig:ResultsForceAndPositionTraining}, $p= 0.003$ and $p= 0.019$, respectively). This performance difference was reduced as the training advanced, with only a (non-significant) 21\% more force error and 10\% more elongation error w.r.t. average participants in the Gaussian group at the end of the training.

\subsection{Exploratory Behavior During Training (H.2)} \label{Sec:ExplorationDuringTraining}

\subsubsection{Overall Exploratory Behavior During Training (H2.1) and the Effect of Spring Type (H2.2)}

Overall, participants' exploratory behavior did not change significantly over training (Fig.~\ref{fig:ResultsExploratory}, pre-release path length: $p= 1$, number of directional changes: $p= 0.524$).

We found distinct exploratory behaviors across spring types (H2.2). In particular, participants who trained with the Gaussian spring exhibited significantly higher \textbf{pre-release path length} and a greater \textbf{number of directional changes} at the beginning of the training compared to those who trained with the Linear spring (Fig.~\ref{fig:ResultsExploratory}, $p< 0.001$ for both). No significant interaction between spring type and trial number was observed for these metrics, indicating that the differences did not significantly change across trials.

\subsubsection{Influence of Personality Traits in Exploration Behavior (H2.3)}
\begin{figure}[h]
    \centering
    \includegraphics[width=\linewidth]{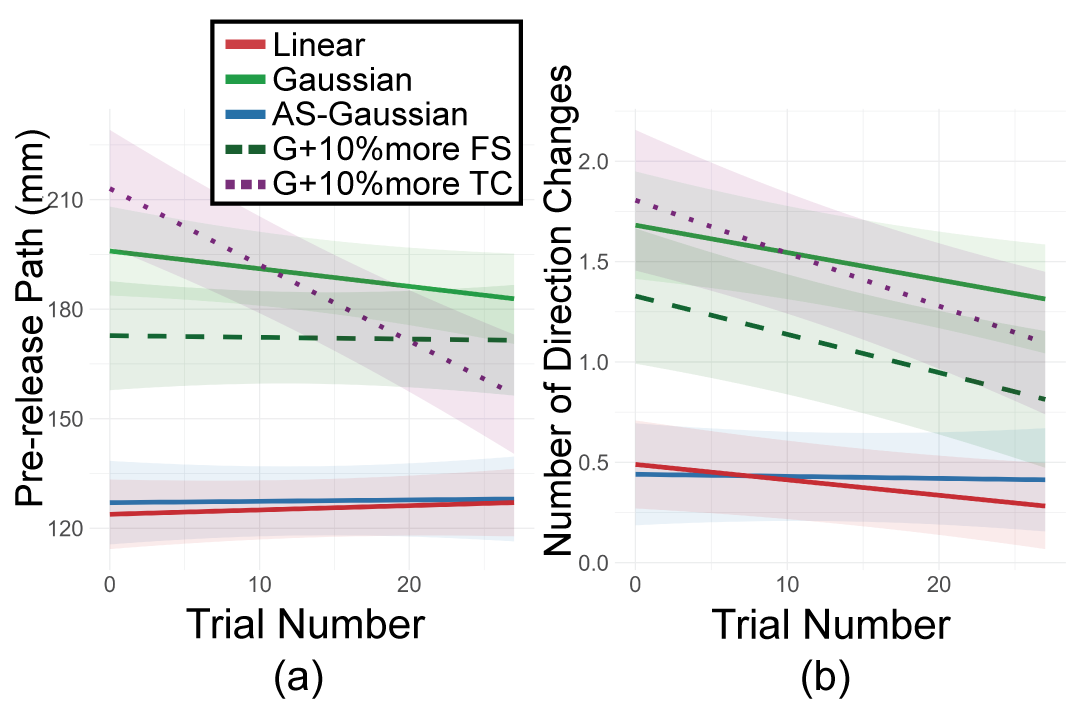}
    \caption{Prediction from the LMM to analyze participants' exploration behaviour during training. Shaded regions indicate 95\% confidence intervals. \textbf{(a)} Evolution of the pre-release path length along the 28 training trials. \textbf{(b)} Evolution of the number of direction changes along the 28 training trials. G: Gaussian, FS: Free Spirit, TC: Transform of Challenge.}
    \label{fig:ResultsExploratory}
\end{figure}

Participants who trained with the Gaussian spring and with higher \textbf{Free Spirit} scores showed significantly lower pre-release path length compared to average participants receiving the same spring 
at the beginning of the training (Fig.~\ref{fig:ResultsExploratory}, $p$ = $0.016$), while the lower number of directional changes did not reach significance after correction ($p = 0.524$). In contrast, participants scoring high in \textbf{Transform of Challenge} and training with the Gaussian spring displayed the opposite tendency, showing increased pre-release path length and number of directional changes at the beginning of the training when compared to average participants allocated to the same group (Fig.~\ref{fig:ResultsExploratory}, $p = 0.003$ and $p = 0.054$). Moreover, a significant interaction between the Transform of Challenge and trial number when receiving the Gaussian spring indicated that these participants reduced both exploration metrics more rapidly over time than others (Fig.~\ref{fig:ResultsExploratory}, pre-release path length: $p < 0.001$, number of directional changes: $p = 0.006$).

\subsection{Motor Learning (H3)}\label{sec:MotorLearning}
\subsubsection{Overall Learning (H3.1) and Effect of Spring Type on Performance Improvements (H3.2)}
\begin{figure}[h]
    \centering
    \includegraphics[width=\linewidth]{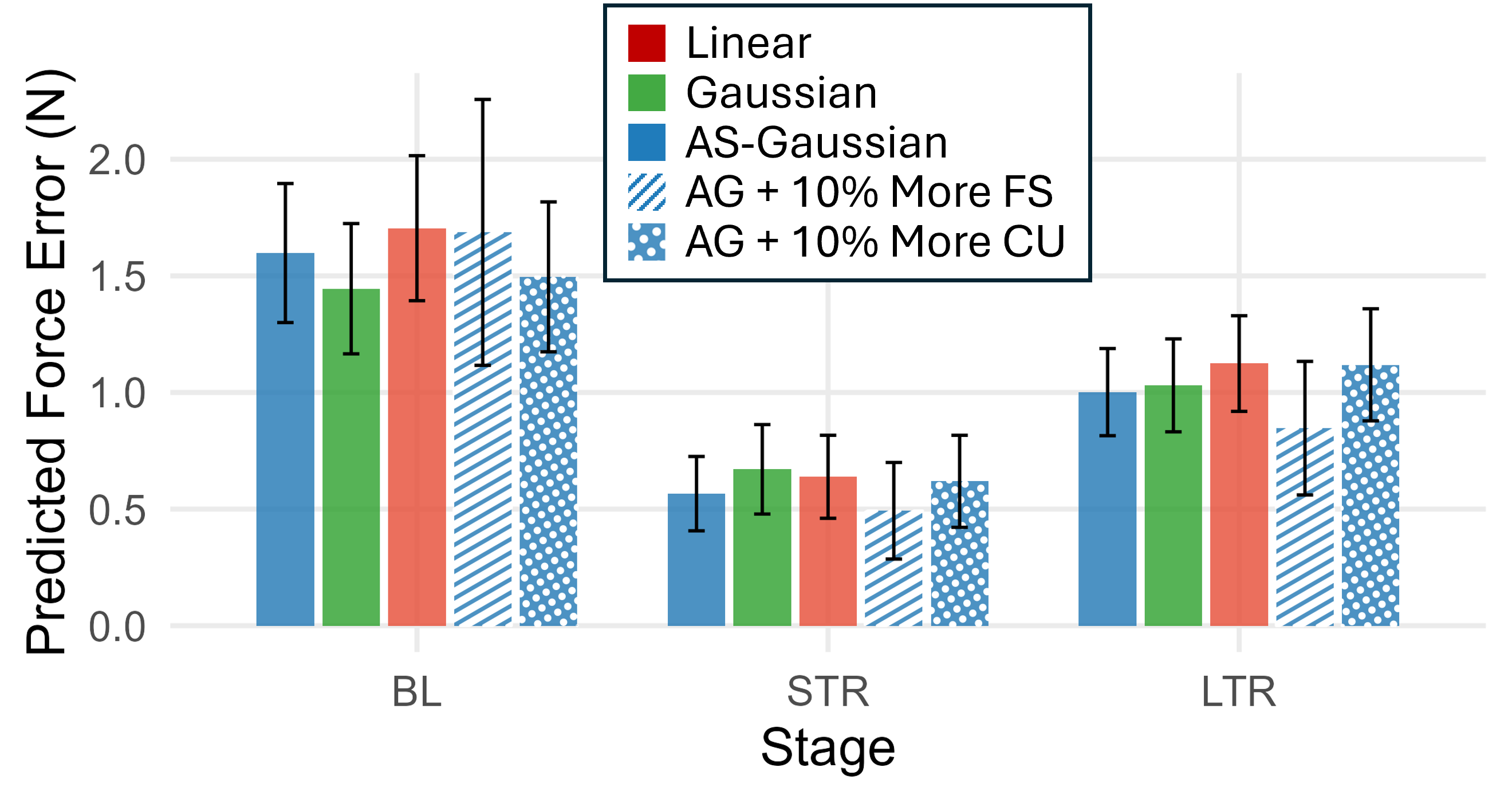}
    \caption{Evolution of the absolute force error along the baseline and the two retention phases. Predicted from the LMM designed to analyze motor learning with 95\% confidence intervals.
    G: Gaussian; AG: Antisymetric Gaussian; FS: Free Spirit; CU: Curiosity}
    \label{fig:ResultsPositionRetention}
\end{figure}

Overall, participants significantly reduced the \textbf{force error} and \textbf{force SD} from Baseline (BL) to Short-term Retention (STR) (Fig.~\ref{fig:ResultsPositionRetention}, $p$ $<$ 0.001 for both) and from BL to Long-term Retention (LTR) ($p = 0.004$ and $p < 0.001$ respectively).
A reduction in both metrics was also observed from the first to the second trial during BL ($p< 0.001$ for both).
However, contrary to our expectations, we did not find a significant effect of spring type on motor learning (H3.2). 

\subsubsection{Influence of Personality Traits in Motor Learning  (H3.3)}
We found interaction effects between the spring condition and personality traits on learning. In particular, individuals allocated in the AS-Gaussian group with high \textbf{Free Spirit} scores achieved a 12\% greater reduction from BL to STR in force error when compared to average participants under the same condition. 
Conversely, participants allocated in the AS-Gaussian group with higher \textbf{Curiosity} scores demonstrated a 9\% smaller reduction from BL to STR in force error compared to average individuals.
However, none of those interactions remained significant after correction (\textit{AS-Gaussian × PhaseSTR × Free Spirit}: $p = 0.076$, \textit{AS-Gaussian × PhaseSTR × Curiosity}: $p= 0.280$). 
We also observed a slightly higher reduction in force SD among high \textbf{Free Spirit} participants allocated to the AS-Gaussian group from BL to STR compared to the average participants in the same group, though this difference was not significant after correction ($p = 0.157$).

\subsection{Transfer Task (H4)}\label{sec:TransferTaskResults}
\begin{figure*}[h]
    \centering
    \includegraphics[width=\linewidth]{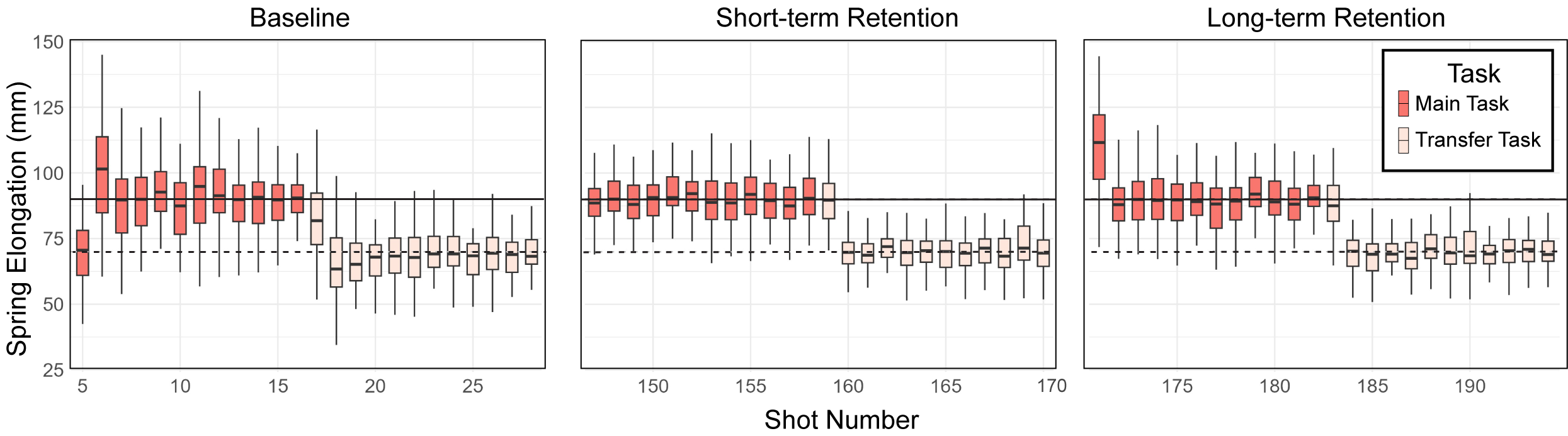}
    \caption{Average spring elongation at release for each shot in BL, STR, and LTR. Colors differentiate between the main and transfer tasks. Black horizontal lines indicate the target position for the main task (90\,mm), and dashed horizontal lines indicate the target position for the transfer task (70\,mm).}
    \label{fig:ResultTransfer}
\end{figure*}

Fig.~\ref{fig:ResultTransfer} shows the average spring elongation at release across participants for each shot along the baseline and retention phases, including both the main and transfer tasks. The results from the LMM showed that participants adjusted the \textbf{spring elongation} at release from an average of 89.96\,mm in the last shot of the main task at baseline (\textit{TrialNumber × ShotNumber}: $p < 0.001$) to 74.83\,mm in the first baseline shot of the transfer task (\textit{TransTask}: $p < 0.001$). Both the $ShotNumber$ and $TrialNumber$, as well as the interaction between them, showed significant effects during the transfer task at baseline, indicating that participants adapted to the new spring dynamics from the first to the subsequent shots and trials
(all $p < 0.001$).

In Fig.~\ref{fig:ResultTransfer} we can also observe that the first shot of the transfer task during both STR and LTR were, on average, closer to the 90\,mm target and the subsequent shots closer to the 70\,mm target, compared to baseline. However, the $TransTask$ variable showed significant interactions with the $Stage$ phase, and its combination with $ShotNumber$, and $TrialNumber$ (all $p < 0.001$) only at LTR. 
Yet, note that the performance degraded at the beginning of the LTR (see first shots in the main task in LTR in Fig.~\ref{fig:ResultTransfer}), and therefore, the main effect of $Stage:LTR$ was large and significant. As a consequence, the interactions with $TransTask$ might reflect how participants corrected the initially larger spring elongation error during the subsequent shots in the LTR phase.

\section{Discussion}\label{sec:Discussion}
\subsection{The Non-Linear Spring Dynamics Modulated Participants' Performance and Exploration Behavior During Training}
Overall, participants improved their performance during training. Yet, this improvement was significant only in reducing elongation error during training, independently of the spring received, while exclusively participants who trained with the Gaussian spring showed significant differences in force error or force SD by the end of training.  
Although this may appear contradictory, in the context of non-linear springs, 
identical spring elongation errors can lead to different force deviations depending on the spring’s force-elongation profile. The force-elongation curve of the AS-Gaussian and the Linear springs intersect at three points: one corresponding to the target force and two additional points (denoted as \textbf{a} and \textbf{b} in Fig.~\ref{fig:SpringProfiles}). 
Within the \textbf{a}–\textbf{b} interval, the AS-Gaussian spring produces forces closer to the target than the Linear spring for equivalent spring elongations. Outside this range, the opposite occurs. 

\begin{figure}[h]
    \centering
    \includegraphics[width=1\linewidth]{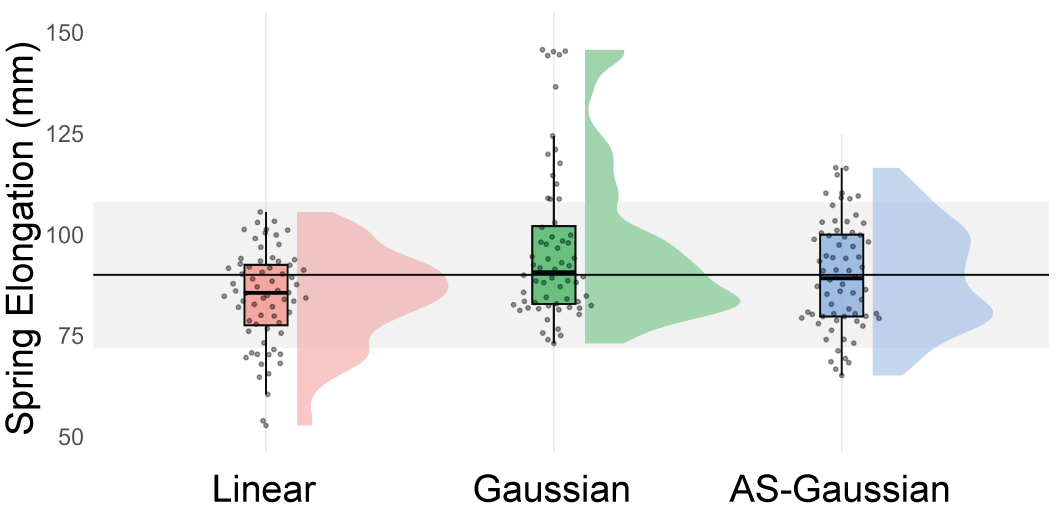}
    \caption{Spring elongations at cube release during the first training trial for each spring condition. Results are presented as raincloud plots, showing the data distribution and jittered individual observations, and are supplemented by boxplots. The black horizontal line marks the target elongation (90\,mm), and the gray band denotes the region bounded by the two intersection points of the Antisymmetric Gaussian and Linear springs, i.e., the \textbf{a}–\textbf{b} interval.}
    \label{fig:Histogram}
\end{figure}

This could be behind the differences in performance across springs, particularly the significantly better force accuracy observed in the AS-Gaussian throughout the training compared to the Linear. However, no significant differences were found between the Gaussian and Linear springs, despite the Gaussian also reducing force error within the \textbf{a}-\textbf{b} interval. The difference between the two non-linear springs may instead reflect how the forces are rendered. 
The AS-Gaussian spring creates a subtle ``wall'' just beyond the target force, due to its steep gradient, resembling performance-enhancing haptic strategies~\cite{Basalp2021}. In contrast, the Gaussian spring reduces the rendered force after the target, which may feel unstable or ``pushing away,'' thereby resembling performance-degrading haptic strategies~\cite{Basalp2021}. Especially early in training, participants might not anticipate this ``pushing'' behavior, making them less likely to remain within the \textbf{a}–\textbf{b} interval and, consequently, less likely to exhibit reasonable force accuracy. As shown in Fig.~\ref{fig:Histogram}, during the first training trial, participants training with the Gaussian spring exhibited 
larger spring elongations at cube release, compared to the other two springs. Yet, as training advanced, the Gaussian group improved their performance, outperforming the Linear one by the end of the training.

Indeed, we hypothesized that the Gaussian spring would challenge participants and thereby improve their performance during training. We also expected that the sudden, noticeable change in spring dynamics, especially the decrease in force after the target force, would prompt them to actively explore the new force–elongation relationship. 
Our results partially support this: participants who trained with the Gaussian spring showed greater exploratory tendencies, as evidenced by greater pre-release path length and number of directional changes, especially during the first training trials, compared to those who trained with the Linear spring. Active exploration has been shown to support motor adaptation when variations directly inform the underlying task structure \cite{Wu2014}, and therefore, we expected this increased exploration behavior to ultimately enhance performance. While we found significant improvements in force accuracy, these were not accompanied by significant reductions in elongation error. It is therefore unclear whether better force accuracy was due to participants releasing the cube in the \textbf{a}-\textbf{b} spring elongation range, as explained before.

\subsection{The Effect of the Spring Dynamics on Performance and Exploratory Behavior Depends on Personality Traits}
Participants’ responses to the spring dynamics modulation likely depended not only on their perception of the haptic rendering but also on what they considered important to achieve within the task. 
In the current task, when the physical interaction with the cube differed from what was expected (i.e., the Gaussian spring), participants who relied on a goal (``high'' Achievers) would not perform better than ``average/low'' Achievers, as they did not know what to do in order to ``achieve higher.'' 
Personality traits like the Transform of Challenge or the Achiever gaming style may require a clearly defined link between actions and goals to activate their motivational drive.
Because the force–elongation relationship of the Gaussian spring was probably not intuitive, these participants may have lacked a concrete sense of progress or accomplishment, preventing their motivational traits from translating into better performance.
This interpretation is supported by the results found regarding the exploratory metrics: participants high in Transform of Challenge scores showed increased exploratory behaviors when allocated to the Gaussian spring, suggesting that they tried hard to master the task, but their efforts did not yield measurable performance gains---likely due to insufficient task clarity or limitations to understand the feedback provided.

\begin{figure}[h]
    \centering
    \includegraphics[width=1\linewidth]{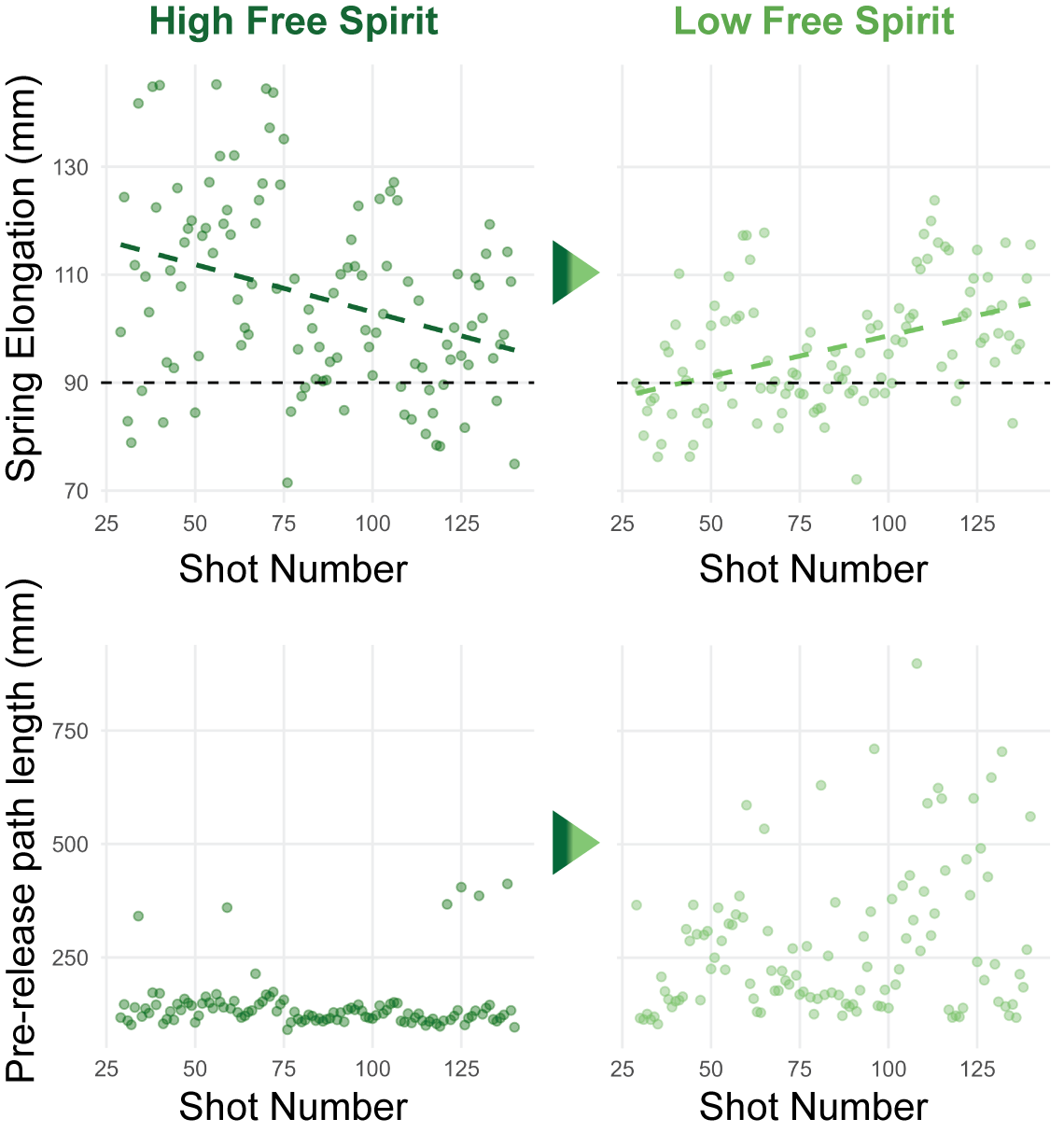}
    \caption{Spring elongation at release per shot (top panels) and pre-release path length per shot (bottom panels) for two representative participants training with the Gaussian spring. The left panels correspond to the participant with the highest Free Spirit score (dark green), and the right panels to the participant with the lowest score (light green). Black dashed horizontal lines indicate the target position, and green dashed lines represent linear regressions illustrating performance changes across training. For High Free Spirit vs. spring elongation: $R^2=0.11$, $p<0.001$. For Low Free Spirit vs. spring elongation; $R^2=0.19$, $p<0.001$.}
    \label{fig:SingleParticipant}
\end{figure}

Free spirits, on the other hand, are intrinsically motivated by exploration, and therefore, we hypothesized that they would increase their exploratory metrics when the interaction invites discovery rather than direct goal achievement.
Looking at the results, one would be surprised by the lower pre-release path length and the absence of significance on the number of directional changes in participants who scored high in Free Spirit when compared to average participants. Yet, it is possible that this exploration was performed \textit{across} consecutive shots rather than \textit{within} a shot. 
That is, rather than systematically exploring the force–elongation relationship within a shot, they might have adopted a ``trial-and-reset'' approach, testing the resulting score of shooting at different elongations. 
Because these participants are typically less goal-driven and more intrinsically motivated to ``try things out''~\cite{Marczewski2015}, they may have tolerated repeated ``failed'' attempts as part of their exploration process. This might be behind the worsened performance observed in this particular group at the beginning of the training. 
We depict this in Fig.~\ref{fig:SingleParticipant}: The upper row shows how a participant with one of the highest Free Spirit scores in our datasets performed across shots during training (left) compared with one of the participants with the lowest scores (right), both in the Gaussian spring group. The lower row shows the evolution of the pre-release path length for each of the exemplary participants.

Together, in conditions where the haptic rendering/guidance is intuitive, users might align their motivations with task objectives, as found in our previous work~\cite{Garzas2024}. However, when the haptic rendering becomes ambiguous---as in the Gaussian spring---differences in exploratory drive and goal orientation begin to dominate performance outcomes and movement behaviors.
Future designs of adaptive haptic systems should therefore consider not only the mechanical properties of feedback but also users' motivational profiles, potentially tailoring haptic strategies to individual tendencies toward exploration or goal pursuit. Furthermore, future directions include exploring how providing information about the system ``surreal'' dynamics would affect performance or task understanding in these particular participant profiles, compared to exploiting implicit learning without task dynamics information~\cite{Bernardoni2019}.

\subsection{The (not Significant) Effect of Spring Dynamics and Personality on Motor Learning}
The few significant differences observed in performance and exploration behavior across springs during training did not result in differences in motor learning between groups. The same can be said of the interaction between spring dynamics and personality traits. In general, 
participants improved in both force and spring elongation accuracy after training, regardless of the spring type used during training. 

The fact that no better performance was found for high Free Spirit participants allocated to the Gaussian spring when comparing baseline with STR is consistent with the training results: during training, although Free Spirit participants allocated to the Gaussian curve improved significantly, many of them probably did not fully grasp the relationship between their actions and task success by the end of the training. 
From the results in Fig.~\ref{fig:ResultsForceAndPositionTraining}, it seems like they were still improving their performance at the end of the training, and therefore, it is an open question whether providing more training time would have resulted in better learning, when compared to average participants receiving the Gaussian spring.

The small differences observed between the baseline and retention phases for participants trained with the AS-Gaussian spring and who had high levels of Free Spirit or Curiosity were unexpected and should be interpreted with caution. 
Given the limited magnitude of those differences, the size of the confidence intervals, and the lack of significance after Holm correction, these effects may reflect statistical noise or incidental variability rather than robust learning differences.
If these trends are not purely incidental, a possible explanation is that the AS-Gaussian profile supported different exploratory strategies depending on individual traits during training. Free Spirit participants may have been more sensitive to the localized increase in stiffness---the ``subtle wall''---and explored behavior around this region, whereas highly curiosity-driven participants may have engaged in broader exploration, including spring elongations around the plateau region. Nevertheless, this extra analysis was considered out of scope for this work; the observed effects were small, with wide confidence intervals, highlighting the potential limitations for drawing firm conclusions. Further work with more targeted measures of exploratory behavior and a larger sample size would be required to evaluate this explanation.

\subsection{Learning the Target Force vs. the Target Elongation}
The transfer task was included to test whether participants could recall and adapt a target force independently of proprioceptive or positional cues, i.e., whether participants learned the target force or the spring elongation at release.  
We found that during the baseline transfer task, participants’ first shot was already shifted toward the transfer target spring elongation. Yet, several more shots were needed to further approach the target elongation of the transfer spring. 
In contrast, during both short- and long-term retention, participants’ first transfer shot tended to remain closer to the previously learned spring elongation (i.e., 90\,mm in the main spring). This was nevertheless followed by a rapid recalibration in the second shot toward the new elongation target. 

This pattern suggests a shift in control strategy after training. During baseline, participants appeared to rely on force-related cues, allowing them to partially adapt to the new force–elongation relationship even on the first exposure to the new spring stiffness. As training progressed, participants likely relied more on muscle memory, reproducing a learned movement pattern rather than recalculating force requirements. Nevertheless, when this strategy failed---in the first transfer shot---participants could rapidly recalibrate their actions, likely re-engaging force-based representations or sensorimotor memory to restore task performance. This aligns with motor learning literature, which suggests that motor learning depends on the history of prior actions, with initial learning characterized by flexible feedback-based adaptation and later performance dominated by context-specific motor representations that generalize only when prediction errors prompt rapid recalibration~\cite{Krakauer2006}.
While we tried to reduce the dependency on proprioceptive cues by including variability in the position of participants with respect to the robot, future work should explore how to better reduce the reliance on muscle memory during training. 

\subsection{Study Limitations}
This work is not exempt from limitations. Apart from the ones mentioned in previous sections, it is worth noting some procedural factors that may have influenced retention outcomes. The repetitive use of the same arm during training led to reported fatigue and shoulder strain, which could have affected both motivation and performance consistency during the first session. 
Muscle fatigue is known to alter proprioceptive acuity, increase motor noise, and reduce fine force control, all of which can interfere with the consolidation and retention of motor learning~\cite{Branscheidt2019}. Future iterations of this study could mitigate these effects by alternating limbs or introducing longer rest periods.
Additionally, as this study was conducted with healthy participants, the generalization of these findings to stroke survivors or other rehabilitation populations remains uncertain and warrants further investigation.

\section{Conclusion}\label{sec:Conclusion}
We investigated how altering the haptically perceived dynamics of virtual objects can enhance learning of precision force generation and whether personality traits modulate this effect. In a parallel design study, fifty participants trained with a haptic device that rendered a virtual spring with either a linear, Gaussian, or Antisymmetric Gaussian force-elongation relationship, the latter two with zero derivative on the target force to be learned. We found that altering the object dynamics can increase task exploration and modulate performance during training. The effect seems to depend on participants' personality traits; 
when task dynamics are not fully understood, goal-oriented individuals may engage in a more systematic and extensive exploration before selecting the target force, whereas high Free Spirit individuals may show lower exploration, but higher shot-to-shot variability in the selected force. Interestingly, these differences did not necessarily translate into improved performance at the end of the training or into better motor learning. We also found that, during the retention phases, participants appeared to rely primarily on the learned target elongation rather than target force, requiring performance errors to shift back to force-based information when task dynamics changed.
Future work should determine whether reducing reliance on proprioceptive information would improve the effectiveness of our method to train precision force generation and clarify how task dynamics, information about the task's goal, and personality traits interact, enabling sensorimotor training to be strategically tailored to specific patient and therapeutic objectives.

\section*{Data Availability}
This article has supplementary downloadable material available at https://XXXXXXXXX/doi, provided by the authors. The dataset used in this study is publicly accessible at XXXXXXX/zenodo.org.

\section*{Acknowledgements}
We sincerely thank Dr. Amy Bellitto for her support during manuscript preparation and Toranj Ahmadniya for her hard work and assistance with the final phase of data collection.

\appendices
\section{\break Model Selection}
Several candidate models were compared, varying in their inclusion of fixed effects, interaction terms, and combinations of personality traits. The naming convention for the following sub-sections in this Appendix matches that of the Statistical Analysis section of the main manuscript to facilitate correspondence between the hypothesis to be answered and its respective linear mixed models used to compare. Model selection was performed through stepwise comparison, guided by the Akaike Information Criterion (AIC) and Bayesian Information Criterion (BIC). When competing models showed similar fit, model selection was further guided by effect size comparisons.

\subsection{LMM to Analyze Performance During Training (H1)}
To compare these LMMs, we employed a dataset including information only from the training phase.
The abbreviations used in the following models refer to the following variables: $SpringType$: Active spring during the corresponding dataframe, i.e., Linear, Gaussian, or AS-Gaussian; $TrialNumber$: an ordinal from 0 to 27 indicating the training trial; $ID$: Participant identifier number; ${FS}_{c}$: Centered and normalized score from the Free Spirit gaming style Questionnaire; ${AC}_{c}$: Centered and normalized score from the Achiever gaming style Questionnaire; ${CH}_{c}$: Centered and normalized score from the Transformation of Challenge sub-scale from the Autotelic Personality Questionnaire;.${BO}_{c}$: Centered and normalized score from the Transformation of Boredom sub-scale from the Autotelic Personality Questionnaire; ${CU}_{c}$: Centered and normalized score from the Curiosity sub-scale from the Autotelic Personality Questionnaire; ${LOC}_{c}$: Centered and normalized score from the Locus of Control Questionnaire.

Model 1:
\begin{align*}
Var_{per} = \,
&SpringType \times TrialNumber\times \\& ({FS}_{c}+{AC}_{c}+{CH}_{c}+{BO}_{c}+{CU}_{c}+{LOC}_{c})\\&
 + (1|ID).
\end{align*}

Model 2:
\begin{align*}
Var_{per} = \,
&SpringType \times TrialNumber \times({FS}_{c}+{CH}_{c})\\&
 + (1|ID).
\end{align*}

Model 3:
\begin{align*}
Var_{per} = \,
&SpringType \times TrialNumber \times({AC}_{c}+{CU}_{c})\\&
 + (1|ID).
\end{align*}


Model 4 (Model chosen for manuscript):
\begin{align*}
Var_{per} = \,
&SpringType \times TrialNumber \times{FS}_{c}
 + (1|ID).
\end{align*}

Model 5:
\begin{align*}
Var_{per} = \,
&SpringType \times TrialNumber \times{AC}_{c}
 + (1|ID).
\end{align*}

Model 6:
\begin{align*}
Var_{per} = \,
&SpringType \times TrialNumber \times{CH}_{c}
 + (1|ID).
\end{align*}

Model 7:
\begin{align*}
Var_{per} = \,
&SpringType \times TrialNumber \times{BO}_{c}
 + (1|ID).
\end{align*}

Model 8:
\begin{align*}
Var_{per} = \,
&SpringType \times TrialNumber \times{CU}_{c}
 + (1|ID).
\end{align*}

Model 9:
\begin{align*}
Var_{per} = \,
&SpringType \times TrialNumber \times{LOC}_{c}
 + (1|ID).
\end{align*}

We compared models 1-9 using AIC and BIC. Results can be found in Table~\ref{AP_A:ModelComparisonTraining}.

Model 4 and Model 7 exhibited the lowest AIC and BIC for both of the dependent variables. 
Because their fit was comparable, model selection was further guided by the magnitude and relevance of the effect sizes of interest. 
After comparing effect sizes across Models 4–9, Model 4 was retained, as it captured the most relevant effects for discussion. The traits included in Models 5, 6, and 8 showed no significant interactions, and the corresponding effects in Model 7 were considered small or non-relevant contributions.

\begin{table}[!h]
\centering
\caption{AIC and BIC values from models 1 to 9. These values were employed to select the best candidate to be the model to analyze performance during training}\label{AP_A:ModelComparisonTraining}
\begin{tabular}{lcccc}
\toprule
& \multicolumn{2}{c}{$\log(|ForceError|)$} & \multicolumn{2}{c}{$\log(|SpringElongationError|)$} \\
\cmidrule(lr){2-3} \cmidrule(lr){4-5}
Model & AIC & BIC & AIC & BIC \\
\midrule
Model 1  & 1050.36 & 1281.10 & -69.66 & 161.08 \\
Model 2  & 963.21 & 1068.09 & -190.20 & -85.31 \\
Model 3  & 978.54 & 1083.43 & -175.80 & -70.91 \\
Model 4  & 930.37 & 1003.79 & -226.45 & -153.03 \\
Model 5  & 951.34 & 1024.76 & -209.25 & -135.83 \\
Model 6  & 956.07 & 1029.49 & -206.76 & -133.34 \\
Model 7  & 925.13 & 998.55 & -230.48 & -157.06 \\
Model 8  & 952.17 & 1025.59 & -208.40 & -134.99 \\
Model 9  & 962.69 & 1036.11 & -157.43 & -124.01 \\
\bottomrule
\end{tabular}
\end{table}

\subsection{LMM to Analyze Participants' Exploration Behavior During Training (H2)}
The dataset and variables employed in the following models are the same as in the previous sub-section, with the only change being the dependent variables, as shown in the main manuscript.

Model 1:
\begin{align*}
Var_{beh} = \,
&SpringType \times TrialNumber \times({FS}_{c})
 + (1|ID).
\end{align*}

Model 2:
\begin{align*}
Var_{beh} = \,
&SpringType \times TrialNumber \times({CH}_{c})
 + (1|ID).
\end{align*}

Model 3:
\begin{align*}
Var_{beh} = \,
&SpringType \times TrialNumber \times({BO}_{c})
 + (1|ID).
\end{align*}

Model 4 (Model chosen for manuscript):
\begin{align*}
Var_{beh} = \,
&SpringType \times TrialNumber \times\\&
({FS}_{c} + {CH}_{c})
 + (1|ID).
\end{align*}

Model 5:
\begin{align*}
Var_{beh} = \,
&SpringType \times TrialNumber \times\\&
({FS}_{c} + {AC}_{c})
 + (1|ID).
\end{align*}

Model 6:
\begin{align*}
Var_{beh} = \,
&SpringType \times TrialNumber \times\\&
({FS}_{c} + {BO}_{c})
 + (1|ID).
\end{align*}

Model 7:
\begin{align*}
Var_{beh} = \,
&SpringType \times TrialNumber \times\\&
({FS}_{c} + {CU}_{c})
 + (1|ID).
\end{align*}

Model 8:
\begin{align*}
Var_{beh} = \,
&SpringType \times TrialNumber \times\\&
({FS}_{c} + {LOC}_{c})
 + (1|ID).
\end{align*}

\medskip

We compared models 1-8 using AIC and BIC. Results can be found in Table~\ref{AP_A:Exploration}.
Given that the combination of Free Spirit and Challenge yielded the best fit for the dependent variable $DistanceTravelled$ and showed comparable fit across models for $DirectionChanges$, we further examined effect sizes across models that included the same traits, i.e., Models 1, 2, and 7. As these models provided overlapping information and very similar AIC and BIC, Model 4 (including both traits) was retained for the main manuscript to ensure simplicity within the analysis while avoiding redundancy across model specifications.

\begin{table}[!h]
\centering
\caption{AIC and BIC values from models 1 to 8. These values were employed to select the best candidate to be the model to analyze exploration during training.}\label{AP_A:Exploration}
\begin{tabular}{lcccc}
\toprule
& \multicolumn{2}{c}{$DistanceTravelled$} & \multicolumn{2}{c}{$DirectionChanges$} \\
\cmidrule(lr){2-3} \cmidrule(lr){4-5}
Model & AIC & BIC & AIC & BIC \\
\midrule
Model 1  & 14374.47 & 14447.89 & 3435.960 & 3509.379 \\
Model 2  & 14318.50 & 14391.92 & 3445.93 & 3519.34 \\
Model 3  & 14393.66 & 14467.08 & 3449.06 & 3522.48 \\
Model 4  & 14274.57 & 14379.46 & 3443.860 & 3548.744 \\
Model 5  & 14343.12 & 14448.00 & 3448.578 & 3553.462 \\
Model 6  & 14344.05 & 14448.93 & 3442.646 & 3547.531 \\
Model 7  & 14338.24 & 14443.12 & 3453.786 & 3558.670 \\
Model 8  & 14358.44 & 14463.32 & 3460.331 & 3565.216 \\
\bottomrule
\end{tabular}
\end{table}

\subsection{LMM to Analyze Motor Learning (H3)}
To compare the LMM to analyze Motor Learning, we employed a dataset excluding information from the training phase and transfer task. The only different variables included w.r.t. the previous sections are: $Condition$: with the group allocated; $Stage$, including the stage, i.e., Baseline, Short-term retention, or Long-term retention. The proposed models were the following.

Model 1:
\begin{align*}
Var_{learn} = \,
&{FS}_{c}\times
(Condition \times Stage + TrialNumber)\\&
+Stage \times TrialNumber
 + (1|ID).
\end{align*}

Model 2:
\begin{align*}
Var_{learn} = \,
&{CU}_{c}\times
(Condition \times Stage + TrialNumber)\\&
+Stage \times TrialNumber
 + (1|ID).
\end{align*}

Model 3:
\begin{align*}
Var_{learn} = \,
&{CH}_{c}\times
(Condition \times Stage + TrialNumber)\\&
+Stage \times TrialNumber
 + (1|ID).
\end{align*}

Model 4:
\begin{align*}
Var_{learn} = \,
&{BO}_{c}\times
(Condition \times Stage + TrialNumber)\\&
+Stage \times TrialNumber
 + (1|ID).
\end{align*}

Model 5:
\begin{align*}
Var_{learn} = \,
&{AC}_{c}\times
(Condition \times Stage + TrialNumber)\\&
+Stage \times TrialNumber
 + (1|ID).
\end{align*}

Model 6:
\begin{align*}
Var_{learn} = \,
&{LOC}_{c}\times
(Condition \times Stage + TrialNumber)\\&
+Stage \times TrialNumber
 + (1|ID).
\end{align*}

Model 7 (Model chosen for manuscript):
\begin{align*}
Var_{learn} = \,
&({FS}_{c} + {CU}_{c})\times \\&
(Condition \times Stage + TrialNumber)\\&
+Stage \times TrialNumber
 + (1|ID).
\end{align*}

Model 8:
\begin{align*}
Var_{learn} = \,
&({FS}_{c} + {CH}_{c})\times \\&
(Condition \times Stage + TrialNumber)\\&
+Stage \times TrialNumber
 + (1|ID).
\end{align*}

Model 9:
\begin{align*}
Var_{learn} = \,
&({FS}_{c} + {BO}_{c})\times \\&
(Condition \times Stage + TrialNumber)\\&
+Stage \times TrialNumber
 + (1|ID).
\end{align*}

Model 10:
\begin{align*}
Var_{learn} = \,
&({FS}_{c} + {AC}_{c})\times \\&
(Condition \times Stage + TrialNumber)\\&
+Stage \times TrialNumber
 + (1|ID).
\end{align*}

We compared models 1-10 using AIC and BIC. Results can be found in Table~\ref{AP_A:MotorLearning}.

In this case, all models showed a very similar fit. When Curiosity and Free Spirit traits were included in the model, only those models showed significant and meaningful effect sizes. When comparing the size effect in Models 1, 2, and 7 (the ones including those traits), they provided overlapping information; therefore, for the sake of simplicity and clearer interpretation, Model 7 was selected for the final analysis, as it provided the most relevant results for the conclusions.

\begin{table}[!h]
\centering
\caption{AIC and BIC values from Models 1--10. These values were used to identify the best candidate model to analyze motor learning.}
\label{AP_A:MotorLearning}
\begin{tabular}{lcc}
\toprule
Model & AIC & BIC \\
\midrule
Model 1  & -73.98038 & 14.91040 \\
Model 2  & -70.74191 & 18.14887 \\
Model 3  & -74.73864 & 14.15214 \\
Model 4  & -71.34454 & 17.54624 \\
Model 5  & -75.73161 & 13.15917 \\
Model 6  & -43.28802 & 45.60276 \\
Model 7  & -69.29499 & 56.63361 \\
Model 8  & -65.31292 & 60.61568 \\
Model 9  & -55.21136 & 70.71724 \\
Model 10 & -65.24330 & 60.68531 \\
\bottomrule
\end{tabular}
\end{table}

\section{Linear Mixed Models output}


\begin{table*}[t]
\centering
\small
\renewcommand{\arraystretch}{1.3}
\begin{tabular}{lrrrrl}
\hline
\textbf{Variable} & \textbf{Estimate} ($\boldsymbol{\beta}$) & \textbf{Std.\ Error} & \textbf{t value} & $\mathbf{p}$\textbf{-value} & \textbf{Corr.\ $p$-value} \\ \hline
(Intercept) & 9.794e-01 & 2.053e-02 & 47.720 & $< 2\times10^{-16}$ & $\mathbf{< 2\times10^{-16}}$\textbf{***} \\
$SpringType$ Gaussian & 8.023e-02 & 3.126e-02 & 2.567 & 0.01040 & 0.09361 \\
$SpringType$ AS-Gaussian & -3.516e-02 & 3.002e-02 & -1.171 & 0.24185 & 1 \\
$TrialNumber$ & -3.044e-03 & 1.007e-03 & -3.024 & 0.00255 & \textbf{0.02546*} \\
$FS_c$ & 2.048e-01 & 1.969e-01 & 1.041 & 0.29912 & 1 \\
$SpringType$ Gaussian : $TrialNumber$ & -3.693e-03 & 1.739e-03 & -2.124 & 0.03383 & 0.2706 \\
$SpringType$ AS-Gaussian : $TrialNumber$ & 1.806e-03 & 1.675e-03 & 1.078 & 0.28114 & 1 \\
$SpringType$ Gaussian : $FS_c$ & 9.095e-01 & 2.888e-01 & 3.149 & 0.00168 & \textbf{0.01852*} \\
$SpringType$ AS-Gaussian : $FS_c$ & -4.263e-01 & 3.283e-01 & -1.298 & 0.19450 & 1 \\
$TrialNumber$ : $FS_c$ & 7.221e-03 & 9.565e-03 & 0.755 & 0.45042 & 1 \\
$SpringType$ Gaussian : $TrialNumber$ : $FS_c$ & -3.220e-02 & 1.608e-02 & -2.002 & 0.04551 & 0.31859 \\
$SpringType$ AS-Gaussian : $TrialNumber$ : $FS_c$ & 2.094e-02 & 1.812e-02 & 1.156 & 0.24804 & 1 \\
\hline
\end{tabular}
\caption{\textbf{Results from the linear mixed-effects model to analyze spring elongation during training.} *($p < 0.05$), **($p < 0.01$), ***($p < 0.001$). Std.: Standard; Corr.: Corrected; FS: Free Spirit. Lowercase 'c' denotes mean-centered questionnaire scores.}
\label{tab:FreeSpiritTrainingModel}
\end{table*}

\begin{table*}[t]
\centering
\small
\renewcommand{\arraystretch}{1.3}
\begin{tabular}{lrrrrl}
\hline
\textbf{Variable} & \textbf{Estimate} ($\boldsymbol{\beta}$) & \textbf{Std.\ Error} & \textbf{t value} & $\mathbf{p}$\textbf{-value} & \textbf{Corr.\ $p$-value} \\ \hline
(Intercept) & 2.514e-02 & 3.080e-02 & 0.816 & 0.415075 & 1 \\
$SpringType$ Gaussian & -3.036e-02 & 4.727e-02 & -0.642 & 0.520911 & 1 \\
$SpringType$ AS-Gaussian & -2.075e-01 & 4.541e-02 & -4.568 & 5.49e-06 & $\mathbf{6.59e-05}$\textbf{***} \\
$TrialNumber$ & -3.044e-03 & 1.529e-03 & -1.991 & 0.046674 & 0.42007 \\
$FS_c$ & 2.065e-01 & 2.952e-01 & 0.699 & 0.484915 & 1 \\
$SpringType$ Gaussian : $TrialNumber$ & -8.662e-03 & 2.640e-03 & -3.281 & 0.001061 & \textbf{0.0106*} \\
$SpringType$ AS-Gaussian : $TrialNumber$ & 6.751e-04 & 2.544e-03 & 0.265 & 0.790767 & 1 \\
$SpringType$ Gaussian : $FS_c$ & 1.584e+00 & 4.369e-01 & 3.627 & 0.000301 & \textbf{0.00331**} \\
$SpringType$ AS-Gaussian : $FS_c$ & -7.242e-01 & 4.964e-01 & -1.459 & 0.144899 & 0.86939 \\
$TrialNumber$ : $FS_c$ & 7.207e-03 & 1.452e-02 & 0.496 & 0.619809 & 1 \\
$SpringType$ Gaussian : $TrialNumber$ : $FS_c$ & -4.234e-02 & 2.442e-02 & -1.734 & 0.083203 & 0.64882 \\
$SpringType$ AS-Gaussian : $TrialNumber$ : $FS_c$ & 4.802e-02 & 2.751e-02 & 1.746 & 0.081103 & 0.64882 \\
\hline
\end{tabular}
\caption{\textbf{Results from the linear mixed-effects model to analyze force error during training.} *($p < 0.05$), **($p < 0.01$), ***($p < 0.001$). Std.: Standard; Corr.: Corrected; FS: Free Spirit. Lowercase 'c' denotes mean-centered questionnaire scores.}
\label{tab:FreeSpiritTrainingModel_DV2}
\end{table*}

\begin{table*}[t]
\centering
\small
\renewcommand{\arraystretch}{1.3}
\begin{tabular}{lrrrrl}
\hline
\textbf{Variable} & \textbf{Estimate} ($\boldsymbol{\beta}$) & \textbf{Std.\ Error} & \textbf{t value} & $\mathbf{p}$\textbf{-value} & \textbf{Corr.\ $p$-value} \\ \hline
(Intercept) & 123.816 & 4.866 & 25.443 & $< 2\times10^{-16}$ & $\mathbf{< 2\times10^{-16}}$\textbf{***} \\
$SpringType$ Gaussian & 72.105 & 6.242 & 11.552 & $< 2\times10^{-16}$ & $\mathbf{< 2\times10^{-16}}$\textbf{***} \\
$SpringType$ AS-Gaussian & 3.182 & 5.865 & 0.542 & 0.587579 & 1 \\
$TrialNumber$ & 0.118 & 0.189 & 0.626 & 0.531608 & 1 \\
$FS_c$ & -34.330 & 47.986 & -0.715 & 0.475890 & 1 \\
$CH_c$ & 3.629 & 40.752 & 0.089 & 0.929162 & 1 \\
$SpringType$ Gaussian : $TrialNumber$ & -0.603 & 0.330 & -1.827 & 0.067930 & 0.88308 \\
$SpringType$ AS-Gaussian : $TrialNumber$ & -0.081 & 0.313 & -0.261 & 0.794388 & 1 \\
$SpringType$ Gaussian : $FS_c$ & -197.508 & 60.503 & -3.264 & 0.001125 & \textbf{0.01576*} \\
$SpringType$ AS-Gaussian : $FS_c$ & 39.333 & 67.584 & 0.582 & 0.560682 & 1 \\
$SpringType$ Gaussian : $CH_c$ & 167.537 & 45.395 & 3.691 & 0.000233 & \textbf{0.00349**} \\
$SpringType$ AS-Gaussian : $CH_c$ & -22.026 & 49.418 & -0.446 & 0.655884 & 1 \\
$TrialNumber$ : $FS_c$ & 0.031 & 1.792 & 0.018 & 0.986001 & 1 \\
$TrialNumber$ : $CH_c$ & -1.115 & 1.672 & -0.667 & 0.504914 & 1 \\
$SpringType$ Gaussian : $TrialNumber$ : $FS_c$ & 4.345 & 3.183 & 1.365 & 0.172487 & 1 \\
$SpringType$ AS-Gaussian : $TrialNumber$ : $FS_c$ & 0.359 & 3.511 & 0.102 & 0.918654 & 1 \\
$SpringType$ Gaussian : $TrialNumber$ : $CH_c$ & -14.912 & 2.479 & -6.015 & $2.31\times10^{-9}$ & $\mathbf{3.7\times10^{-8}}$\textbf{***} \\
$SpringType$ AS-Gaussian : $TrialNumber$ : $CH_c$ & 1.422 & 2.659 & 0.535 & 0.592942 & 1 \\
\hline
\end{tabular}
\caption{\textbf{Results from the linear mixed-effects model to analyze Distance traveled during training.} *($p < 0.05$), **($p < 0.01$), ***($p < 0.001$). Std.: Standard; Corr.: Corrected; FS: Free Spirit; CH: Transform of Challenge. Lowercase `c' denotes mean-centered questionnaire scores.}
\label{tab:FreeSpiritChallengeTrainingModel}
\end{table*}

\begin{table*}[t]
\centering
\small
\renewcommand{\arraystretch}{1.3}
\begin{tabular}{lrrrrl}
\hline
\textbf{Variable} & \textbf{Estimate} ($\boldsymbol{\beta}$) & \textbf{Std.\ Error} & \textbf{t value} & $\mathbf{p}$\textbf{-value} & \textbf{Corr.\ $p$-value} \\ \hline
(Intercept) & 0.490 & 0.112 & 4.372 & $3.24\times10^{-5}$ & $\mathbf{3.5\times10^{-4}}$\textbf{***} \\
$SpringType$ G & 1.192 & 0.125 & 9.539 & $< 2\times10^{-16}$ & $\mathbf{< 2\times10^{-16}}$\textbf{***} \\
$SpringType$ AG & -0.049 & 0.117 & -0.419 & 0.67544 & 1 \\
$TrialNumber$ & -0.008 & 0.004 & -2.059 & 0.03973 & 0.5243 \\
$FS_c$ & -1.008 & 1.113 & -0.905 & 0.36802 & 1 \\
$CH_c$ & -1.395 & 0.927 & -1.504 & 0.13550 & 1 \\
$SpringType$ Gaussian : $TrialNumber$ & -0.006 & 0.007 & -0.914 & 0.36101 & 1 \\
$SpringType$ AS-Gaussian : $TrialNumber$ & 0.007 & 0.006 & 1.080 & 0.28040 & 1 \\
$SpringType$ Gaussian : $FS_c$ & -2.523 & 1.212 & -2.083 & 0.03745 & 0.5243 \\
$SpringType$ AS-Gaussian : $FS_c$ & 0.599 & 1.356 & 0.442 & 0.65879 & 1 \\
$SpringType$ Gaussian : $CH_c$ & 2.638 & 0.906 & 2.914 & 0.00363 & $\mathbf{0.05441}^{.}$ \\
$SpringType$ AS-Gaussian : $CH_c$ & 0.924 & 0.988 & 0.935 & 0.34981 & 1 \\
$TrialNumber$ : $FS_c$ & 0.026 & 0.035 & 0.749 & 0.45424 & 1 \\
$TrialNumber$ : $CH_c$ & 0.046 & 0.033 & 1.406 & 0.15995 & 1 \\
$SpringType$ Gaussian : $TrialNumber$ : $FS_c$ & -0.081 & 0.063 & -1.291 & 0.19687 & 1 \\
$SpringType$ AS-Gaussian : $TrialNumber$ : $FS_c$ & -0.015 & 0.069 & -0.218 & 0.82732 & 1 \\
$SpringType$ Gaussian : $TrialNumber$ : $CH_c$ & -0.174 & 0.049 & -3.556 & 0.00039 & \textbf{0.00624**} \\
$SpringType$ AS-Gaussian : $TrialNumber$ : $CH_c$ & -0.050 & 0.052 & -0.945 & 0.34474 & 1 \\
\hline
\end{tabular}
\caption{\textbf{Results from the linear mixed-effects model to analyze Directional Changes during training.} *($p < 0.05$), **($p < 0.01$), ***($p < 0.001$).}
\label{tab:DirectionChanges}
\end{table*}

\begin{table*}[t]
\centering
\small
\renewcommand{\arraystretch}{1.3}
\begin{tabular}{lrrrrl}
\hline
\textbf{Variable} & \textbf{Estimate} ($\boldsymbol{\beta}$) & \textbf{Std.\ Error} & \textbf{t value} & $\mathbf{p}$\textbf{-value} & \textbf{Corr.\ $p$-value} \\ \hline
(Intercept) & 0.232 & 0.040 & 5.729 & $5.96\times10^{-8}$ & \textbf{1.79e-06***} \\
$FS_c$ & -0.534 & 0.380 & -1.404 & 0.16278 & 1 \\
$CU_c$ & 0.623 & 0.297 & 2.096 & 0.03818 & 0.7255 \\
$Condition$ Gaussian & -0.072 & 0.054 & -1.326 & 0.18773 & 1 \\
$Condition$ AS-Gaussian & -0.028 & 0.053 & -0.530 & 0.59727 & 1 \\
$Stage$ STR & -0.426 & 0.066 & -6.451 & $6.64\times10^{-10}$ & \textbf{2.12e-08***} \\
$Stage$ LTR & -0.181 & 0.047 & -3.865 & 0.000145 & \textbf{0.00406**} \\
$TrialNumber$ & -0.196 & 0.033 & -5.938 & $1.07\times10^{-8}$ & \textbf{3.32e-07***} \\
$Condition$ Gaussian : $Stage$ STR & 0.093 & 0.059 & 1.565 & 0.119028 & 1 \\
$Condition$ AS-Gaussian : $Stage$ STR & -0.024 & 0.058 & -0.416 & 0.677450 & 1 \\
$Condition$ Gaussian : $Stage$ LTR & 0.034 & 0.059 & 0.572 & 0.567806 & 1 \\
$Condition$ AS-Gaussian : $Stage$ LTR & -0.022 & 0.058 & -0.380 & 0.704502 & 1 \\
$\mathbf{FS_c}$ : $Condition$ Gaussian & 1.343 & 0.554 & 2.425 & 0.016995 & 0.3569 \\
$\mathbf{FS_c}$ : $Condition$ AS-Gaussian & 0.768 & 0.658 & 1.168 & 0.245425 & 1 \\
$\mathbf{FS_c}$ : $Stage$ STR & 1.328 & 0.454 & 2.924 & 0.003806 & 0.09134 \\
$\mathbf{FS_c}$ : $Stage$ LTR & 0.351 & 0.401 & 0.875 & 0.382435 & 1 \\
$\mathbf{FS_c}$ : $TrialNumber$ & -0.222 & 0.214 & -1.039 & 0.299804 & 1 \\
$\mathbf{CU_c}$ : $Condition$ Gaussian & -1.479 & 0.471 & -3.139 & 0.002198 & 0.05935 \\
$\mathbf{CU_c}$ : $Condition$ AS-Gaussian & -0.910 & 0.450 & -2.024 & 0.045476 & 0.81858 \\
$\mathbf{CU_c}$ : $Stage$ STR & -0.573 & 0.354 & -1.618 & 0.106956 & 1 \\
$\mathbf{CU_c}$ : $Stage$ LTR & -0.753 & 0.314 & -2.397 & 0.017362 & 0.3569 \\
$\mathbf{CU_c}$ : $TrialNumber$ & 0.152 & 0.164 & 0.923 & 0.356944 & 1 \\
$Stage$ STR : $TrialNumber$ & 0.259 & 0.047 & 5.545 & $8.13\times10^{-8}$ & \textbf{2.36e-06***} \\
$Stage$ LTR : $TrialNumber$ & 0.063 & 0.047 & 1.340 & 0.181441 & 1 \\
$\mathbf{FS_c}$ : $Condition$ Gaussian : $Stage$ STR & -1.478 & 0.608 & -2.430 & 0.015880 & \textbf{0.34936*} \\
$\mathbf{FS_c}$ : $Condition$ AS-Gaussian : $Stage$ STR & -2.165 & 0.722 & -2.997 & 0.003026 & \textbf{0.07564**} \\
$\mathbf{FS_c}$ : $Condition$ Gaussian : $Stage$ LTR & -0.243 & 0.608 & -0.400 & 0.689477 & 1 \\
$\mathbf{FS_c}$ : $Condition$ AS-Gaussian : $Stage$ LTR & -1.313 & 0.722 & -1.818 & 0.070405 & 1 \\
$\mathbf{CU_c}$ : $Condition$ Gaussian : $Stage$ STR & 0.479 & 0.518 & 0.926 & 0.355358 & 1 \\
$\mathbf{CU_c}$ : $Condition$ AS-Gaussian : $Stage$ STR & 1.248 & 0.494 & 2.527 & 0.012189 & 0.28034 \\
$\mathbf{CU_c}$ : $Condition$ Gaussian : $Stage$ LTR & 0.874 & 0.518 & 1.690 & 0.092490 & 1 \\
$\mathbf{CU_c}$ : $Condition$ AS-Gaussian : $Stage$ LTR & 1.519 & 0.494 & 3.076 & 0.002359 & 0.06134 \\
\hline
\end{tabular}
\caption{\textbf{Results from the linear mixed-effects model to analyze motor learning.}}
\label{tab:MotorLearning}
\end{table*}

\begin{table*}[t]
\centering
\small
\renewcommand{\arraystretch}{1.3}
\begin{tabular}{lrrrrl}
\hline
\textbf{Variable} & \textbf{Estimate} ($\boldsymbol{\beta}$) & \textbf{Std.\ Error} & \textbf{t value} & $\mathbf{p}$\textbf{-value} & \textbf{Corr.\ $p$-value} \\ \hline
(Intercept) & 82.255 & 1.274 & 64.579 & $< 2\times10^{-16}$ & $\mathbf{< 2\times10^{-16}}$\textbf{***} \\
$Task$ Tr & -7.425 & 1.768 & -4.199 & $2.74\times10^{-5}$ & \textbf{2.47e-04***} \\
$Stage$ STR & 6.824 & 1.768 & 3.859 & 0.000116 & \textbf{9.26e-04***} \\
$Stage$ LTR & 17.495 & 1.768 & 9.895 & $< 2\times10^{-16}$ & $\mathbf{< 2\times10^{-16}}$\textbf{***} \\
$TrialNumber$ & 9.169 & 1.768 & 5.185 & $2.28\times10^{-7}$ & \textbf{2.96e-06***} \\
$ShotNumber$ & 2.277 & 0.413 & 5.515 & $3.75\times10^{-8}$ & \textbf{5.62e-07***} \\
$Task$ Tr : $Stage$ STR & -1.800 & 2.501 & -0.720 & 0.471801 & 0.5155 \\
$Task$ Tr : $Stage$ LTR & -13.949 & 2.501 & -5.578 & $2.61\times10^{-8}$ & \textbf{4.18e-07***} \\
$Task$ Tr : $TrialNumber$ & -15.019 & 2.504 & -5.998 & $2.20\times10^{-9}$ & \textbf{3.96e-08***} \\
$Stage$ STR : $TrialNumber$ & -8.752 & 2.501 & -3.500 & 0.000471 & \textbf{0.0033**} \\
$Stage$ LTR : $TrialNumber$ & -19.577 & 2.501 & -7.829 & $6.46\times10^{-15}$ & \textbf{1.36e-13***} \\
$Task$ Tr : $ShotNumber$ & -4.162 & 0.584 & -7.126 & $1.24\times10^{-12}$ & \textbf{2.49e-11***} \\
$Stage$ STR : $ShotNumber$ & -1.661 & 0.584 & -2.843 & 0.004490 & \textbf{0.02245*} \\
$Stage$ LTR : $ShotNumber$ & -5.013 & 0.584 & -8.583 & $< 2\times10^{-16}$ & \textbf{3e-16***} \\
$TrialNumber$ : $ShotNumber$ & -2.966 & 0.584 & -5.078 & $4.01\times10^{-7}$ & \textbf{4.81e-06***} \\
$Task$ Tr : $Stage$ STR : $TrialNumber$ & 4.836 & 3.539 & 1.367 & 0.171834 & 0.5155 \\
$Task$ Tr : $Stage$ LTR : $TrialNumber$ & 16.662 & 3.539 & 4.709 & $2.59\times10^{-6}$ & \textbf{2.85e-05***} \\
$Task$ Tr : $Stage$ STR : $ShotNumber$ & 0.936 & 0.826 & 1.133 & 0.257208 & 0.5155 \\
$Task$ Tr : $Stage$ LTR : $ShotNumber$ & 4.497 & 0.826 & 5.445 & $5.52\times10^{-8}$ & \textbf{7.73e-07***} \\
$Task$ Tr : $TrialNumber$ : $ShotNumber$ & 4.787 & 0.827 & 5.791 & $7.60\times10^{-9}$ & \textbf{1.29e-07***} \\
$Stage$ STR : $TrialNumber$ : $ShotNumber$ & 2.620 & 0.826 & 3.172 & 0.001526 & \textbf{9.16e-03**} \\
$Stage$ LTR : $TrialNumber$ : $ShotNumber$ & 5.809 & 0.826 & 7.034 & $2.40\times10^{-12}$ & \textbf{4.56e-11***} \\
$Task$ Tr : $Stage$ STR : $TrialNumber$ : $ShotNumber$ & -1.809 & 1.169 & -1.548 & 0.121770 & 0.48708 \\
$Task$ Tr : $Stage$ LTR : $TrialNumber$ : $ShotNumber$ & -5.163 & 1.169 & -4.418 & $1.02\times10^{-5}$ & \textbf{0.0001***} \\
\hline
\end{tabular}
\caption{\textbf{Results from the linear mixed-effects model to analyze the transfer task during the retention phases.}}
\label{tab:TransferTask}
\end{table*}

\clearpage

\bibliographystyle{IEEEtran}
\balance
\bibliography{references}

\begin{thebibliography}{10}
\providecommand{\url}[1]{#1}
\csname url@samestyle\endcsname
\providecommand{\newblock}{\relax}
\providecommand{\bibinfo}[2]{#2}
\providecommand{\BIBentrySTDinterwordspacing}{\spaceskip=0pt\relax}
\providecommand{\BIBentryALTinterwordstretchfactor}{4}
\providecommand{\BIBentryALTinterwordspacing}{\spaceskip=\fontdimen2\font plus
\BIBentryALTinterwordstretchfactor\fontdimen3\font minus \fontdimen4\font\relax}
\providecommand{\BIBforeignlanguage}[2]{{%
\expandafter\ifx\csname l@#1\endcsname\relax
\typeout{** WARNING: IEEEtran.bst: No hyphenation pattern has been}%
\typeout{** loaded for the language `#1'. Using the pattern for}%
\typeout{** the default language instead.}%
\else
\language=\csname l@#1\endcsname
\fi
#2}}
\providecommand{\BIBdecl}{\relax}
\BIBdecl

\bibitem{Feigin2024}
V.~L. Feigin \emph{et~al.}, ``Global, regional, and national burden of stroke and its risk factors, 1990--2021: a systematic analysis for the global burden of disease study 2021,'' \emph{The Lancet Neurology}, vol.~23, no.~10, pp. 973--1003, 2024.

\bibitem{Connell2008}
L.~A. Connell, N.~Lincoln, and K.~Radford, ``Somatosensory impairment after stroke: frequency of different deficits and their recovery,'' \emph{Clinical rehabilitation}, vol.~22, no.~8, pp. 758--767, 2008.

\bibitem{Kang2015}
N.~Kang and J.~H. Cauraugh, ``Force control in chronic stroke,'' \emph{Neuroscience \& Biobehavioral Reviews}, vol.~52, pp. 38--48, 2015.

\bibitem{Quaney2005}
B.~M. Quaney, S.~Perera, R.~Maletsky, C.~W. Luchies, and R.~J. Nudo, ``Impaired grip force modulation in the ipsilesional hand after unilateral middle cerebral artery stroke,'' \emph{Neurorehabilitation and neural repair}, vol.~19, no.~4, pp. 338--349, 2005.

\bibitem{Lodha2010}
N.~Lodha, S.~K. Naik, S.~A. Coombes, and J.~H. Cauraugh, ``Force control and degree of motor impairments in chronic stroke,'' \emph{Clinical Neurophysiology}, vol. 121, no.~11, pp. 1952--1961, 2010.

\bibitem{Bolognini2016}
N.~Bolognini, C.~Russo, and D.~J. Edwards, ``The sensory side of post-stroke motor rehabilitation,'' \emph{Restorative neurology and neuroscience}, vol.~34, no.~4, pp. 571--586, 2016.

\bibitem{Gassert2018}
R.~Gassert and V.~Dietz, ``Rehabilitation robots for the treatment of sensorimotor deficits: a neurophysiological perspective,'' \emph{Journal of neuroengineering and rehabilitation}, vol.~15, no.~1, p.~46, 2018.

\bibitem{Lee2024}
J.~H. Lee, H.~Lee, H.~Kim, R.-K. Kim, T.~L. Lee, D.-K. Ko, H.~Lee, and N.~Kang, ``Resistance band training with functional electrical stimulation improves force control capabilities in older adults: a preliminary study,'' \emph{EXCLI journal}, vol.~23, p. 130, 2024.

\bibitem{Mcdonnell2007}
M.~N. McDonnell, S.~L. Hillier, T.~S. Miles, P.~D. Thompson, and M.~C. Ridding, ``Influence of combined afferent stimulation and task-specific training following stroke: a pilot randomized controlled trial,'' \emph{Neurorehabilitation and neural repair}, vol.~21, no.~5, pp. 435--443, 2007.

\bibitem{Villar2024}
E.~Villar~Ortega, K.~A. Buetler, E.~A. Aks{\"o}z, and L.~Marchal-Crespo, ``Enhancing touch sensibility with sensory electrical stimulation and sensory retraining,'' \emph{Journal of NeuroEngineering and Rehabilitation}, vol.~21, no.~1, p.~79, 2024.

\bibitem{Dafotakis2008}
M.~Dafotakis, C.~Grefkes, S.~B. Eickhoff, H.~Karbe, G.~R. Fink, and D.~A. Nowak, ``Effects of rtms on grip force control following subcortical stroke,'' \emph{Experimental neurology}, vol. 211, no.~2, pp. 407--412, 2008.

\bibitem{Stefan2008}
K.~Stefan, R.~Gentner, D.~Zeller, S.~Dang, and J.~Classen, ``Theta-burst stimulation: remote physiological and local behavioral after-effects,'' \emph{Neuroimage}, vol.~40, no.~1, pp. 265--274, 2008.

\bibitem{Voelcker2010}
C.~Voelcker-Rehage and B.~Godde, ``High frequency sensory stimulation improves tactile but not motor performance in older adults,'' \emph{Motor control}, vol.~14, no.~4, pp. 460--477, 2010.

\bibitem{Krakauer2019}
J.~W. Krakauer, A.~M. Hadjiosif, J.~Xu, A.~L. Wong, and A.~M. Haith, ``Motor learning,'' \emph{Comprehensive physiology}, vol.~9, no.~2, pp. 613--663, 2019.

\bibitem{Quaney2010}
B.~M. Quaney, J.~He, G.~Timberlake, K.~Dodd, and C.~Carr, ``Visuomotor training improves stroke-related ipsilesional upper extremity impairments,'' \emph{Neurorehabilitation and neural repair}, vol.~24, no.~1, pp. 52--61, 2010.

\bibitem{Kurillo2004}
G.~Kurillo, A.~Zupan, and T.~Bajd, ``Force tracking system for the assessment of grip force control in patients with neuromuscular diseases,'' \emph{Clinical Biomechanics}, vol.~19, no.~10, pp. 1014--1021, 2004.

\bibitem{Guo2022}
J.~Guo, T.~Liu, and J.~Wang, ``Effects of auditory feedback on fine motor output and corticomuscular coherence during a unilateral finger pinch task,'' \emph{Frontiers in Neuroscience}, vol.~16, p. 896933, 2022.

\bibitem{Hsu2012}
H.-Y. Hsu, C.-F. Lin, F.-C. Su, H.-T. Kuo, H.-Y. Chiu, and L.-C. Kuo, ``Clinical application of computerized evaluation and re-education biofeedback prototype for sensorimotor control of the hand in stroke patients,'' \emph{Journal of neuroengineering and rehabilitation}, vol.~9, no.~1, p.~26, 2012.

\bibitem{Lum2004}
P.~S. Lum, C.~G. Burgar, and P.~C. Shor, ``Evidence for improved muscle activation patterns after retraining of reaching movements with the mime robotic system in subjects with post-stroke hemiparesis,'' \emph{IEEE Transactions on Neural Systems and Rehabilitation Engineering}, vol.~12, no.~2, pp. 186--194, 2004.

\bibitem{Fischer2006}
H.~Fischer, L.~Kahn, E.~Pelosin, H.~Roth, J.~Barbas, W.~Z. Rymer, and D.~Reinkensmeyer, ``Can robot-assisted therapy promote generalization of motor learning following stroke?: preliminary results,'' in \emph{The First IEEE/RAS-EMBS International Conference on Biomedical Robotics and Biomechatronics, 2006. BioRob 2006.}\hskip 1em plus 0.5em minus 0.4em\relax IEEE, 2006, pp. 865--868.

\bibitem{Emken2005}
J.~L. Emken and D.~J. Reinkensmeyer, ``Robot-enhanced motor learning: accelerating internal model formation during locomotion by transient dynamic amplification,'' \emph{IEEE Transactions on Neural Systems and Rehabilitation Engineering}, vol.~13, no.~1, pp. 33--39, 2005.

\bibitem{Patton2004}
J.~L. Patton and F.~A. Mussa-Ivaldi, ``Robot-assisted adaptive training: custom force fields for teaching movement patterns,'' \emph{IEEE Transactions on Biomedical Engineering}, vol.~51, no.~4, pp. 636--646, 2004.

\bibitem{Patton2006}
J.~L. Patton, M.~E. Stoykov, M.~Kovic, and F.~A. Mussa-Ivaldi, ``Evaluation of robotic training forces that either enhance or reduce error in chronic hemiparetic stroke survivors,'' \emph{Experimental brain research}, vol. 168, no.~3, pp. 368--383, 2006.

\bibitem{Rauter2011}
G.~Rauter, R.~Sigrist, L.~Marchal-Crespo, H.~Vallery, R.~Riener, and P.~Wolf, ``Assistance or challenge? filling a gap in user-cooperative control,'' in \emph{2011 IEEE/RSJ International Conference on Intelligent Robots and Systems}, 2011, pp. 3068--3073.

\bibitem{Shadmehr1994}
R.~Shadmehr and F.~A. Mussa-Ivaldi, ``Adaptive representation of dynamics during learning of a motor task,'' \emph{Journal of neuroscience}, vol.~14, no.~5, pp. 3208--3224, 1994.

\bibitem{Ratz2024}
R.~R{\"a}tz, F.~Conti, I.~Thaler, R.~M. M{\"u}ri, and L.~Marchal-Crespo, ``Enhancing stroke rehabilitation with whole-hand haptic rendering: development and clinical usability evaluation of a novel upper-limb rehabilitation device,'' \emph{Journal of NeuroEngineering and Rehabilitation}, vol.~21, no.~1, p. 172, 2024.

\bibitem{Basalp2021}
E.~Basalp, P.~Wolf, and L.~Marchal-Crespo, ``Haptic training: which types facilitate (re) learning of which motor task and for whom? answers by a review,'' \emph{IEEE transactions on haptics}, vol.~14, no.~4, pp. 722--739, 2021.

\bibitem{Salisbury2004}
K.~Salisbury, F.~Conti, and F.~Barbagli, ``Haptic rendering: introductory concepts,'' \emph{IEEE computer graphics and applications}, vol.~24, no.~2, pp. 24--32, 2004.

\bibitem{Ritter2023}
C.~Ritter, M.~Senne, N.~Berberich, K.~Yilmazer, N.~Paredes-Acu{\~n}a, and G.~Cheng, ``Grip force dynamics during exoskeleton-assisted and virtual grasping,'' in \emph{2023 International Conference on Rehabilitation Robotics (ICORR)}.\hskip 1em plus 0.5em minus 0.4em\relax IEEE, 2023, pp. 1--6.

\bibitem{Huang2012}
F.~C. Huang and J.~L. Patton, ``Augmented dynamics and motor exploration as training for stroke,'' \emph{IEEE Transactions on Biomedical Engineering}, vol.~60, no.~3, pp. 838--844, 2012.

\bibitem{Ozen2021}
{\"O}.~{\"O}zen, K.~A. Buetler, and L.~Marchal-Crespo, ``Promoting motor variability during robotic assistance enhances motor learning of dynamic tasks,'' \emph{Frontiers in neuroscience}, vol.~14, p. 600059, 2021.

\bibitem{Marchal-Crespo2009}
L.~Marchal-Crespo and D.~J. Reinkensmeyer, ``Review of control strategies for robotic movement training after neurologic injury,'' \emph{Journal of neuroengineering and rehabilitation}, vol.~6, no.~1, pp. 1--15, 2009.

\bibitem{Garzas2025}
A.~Garz{\'a}s-Villar, C.~Boersma, A.~Derumigny, J.~M. Prendergast, A.~Zgonnikov, J.~M. Cramm, and L.~Marchal-Crespo, ``The interplay between haptic guidance and personality traits in robotic-assisted motor learning,'' \emph{Journal of NeuroEngineering and Rehabilitation}, vol.~22, no.~1, p. 238, 2025.

\bibitem{Garzas2024}
A.~Garz{\'a}s-Villar, C.~Boersma, A.~Derumigny, A.~Zgonnikov, and L.~Marchal-Crespo, ``Personality traits modulate the effect of haptic guidance during robotic-assisted motor training,'' in \emph{2024 10th IEEE RAS/EMBS International Conference for Biomedical Robotics and Biomechatronics (BioRob)}.\hskip 1em plus 0.5em minus 0.4em\relax IEEE, 2024, pp. 1023--1028.

\bibitem{Tse2020}
D.~C.~K. Tse, V.~W. Lau, R.~Perlman, and M.~McLaughlin, ``The development and validation of the autotelic personality questionnaire,'' \emph{Journal of personality assessment}, vol. 102, no.~1, pp. 88--101, 2020.

\bibitem{Marczewski2015}
A.~Marczewski, ``Even ninja monkeys like to play,'' \emph{London: Blurb Inc}, vol.~1, no.~1, p.~28, 2015.

\bibitem{Rotter1966}
J.~B. Rotter, ``Generalized expectancies for internal versus external control of reinforcement.'' \emph{Psychological monographs: General and applied}, vol.~80, no.~1, p.~1, 1966.

\bibitem{Piryankova2014}
I.~V. Piryankova, H.~Y. Wong, S.~A. Linkenauger, C.~Stinson, M.~R. Longo, H.~H. B{\"u}lthoff, and B.~J. Mohler, ``Owning an overweight or underweight body: distinguishing the physical, experienced and virtual body,'' \emph{PloS one}, vol.~9, no.~8, p. e103428, 2014.

\bibitem{Wu2014}
H.~G. Wu, Y.~R. Miyamoto, L.~N.~G. Castro, B.~P. {\"O}lveczky, and M.~A. Smith, ``Temporal structure of motor variability is dynamically regulated and predicts motor learning ability,'' \emph{Nature neuroscience}, vol.~17, no.~2, pp. 312--321, 2014.

\bibitem{Bernardoni2019}
F.~Bernardoni, {\"O}.~{\"O}zen, K.~Buetler, and L.~Marchal-Crespo, ``Virtual reality environments and haptic strategies to enhance implicit learning and motivation in robot-assisted training,'' in \emph{2019 IEEE 16th International Conference on Rehabilitation Robotics (ICORR)}, 2019, pp. 760--765.

\bibitem{Krakauer2006}
J.~W. Krakauer, P.~Mazzoni, A.~Ghazizadeh, R.~Ravindran, and R.~Shadmehr, ``Generalization of motor learning depends on the history of prior action,'' \emph{PLoS biology}, vol.~4, no.~10, p. e316, 2006.

\bibitem{Branscheidt2019}
M.~Branscheidt, P.~Kassavetis, M.~Anaya, D.~Rogers, H.~D. Huang, M.~A. Lindquist, and P.~Celnik, ``Fatigue induces long-lasting detrimental changes in motor-skill learning,'' \emph{Elife}, vol.~8, p. e40578, 2019.

\end{thebibliography}

\nobalance

\begin{IEEEbiography}[{\includegraphics[width=1in,height=1.25in,clip,keepaspectratio]{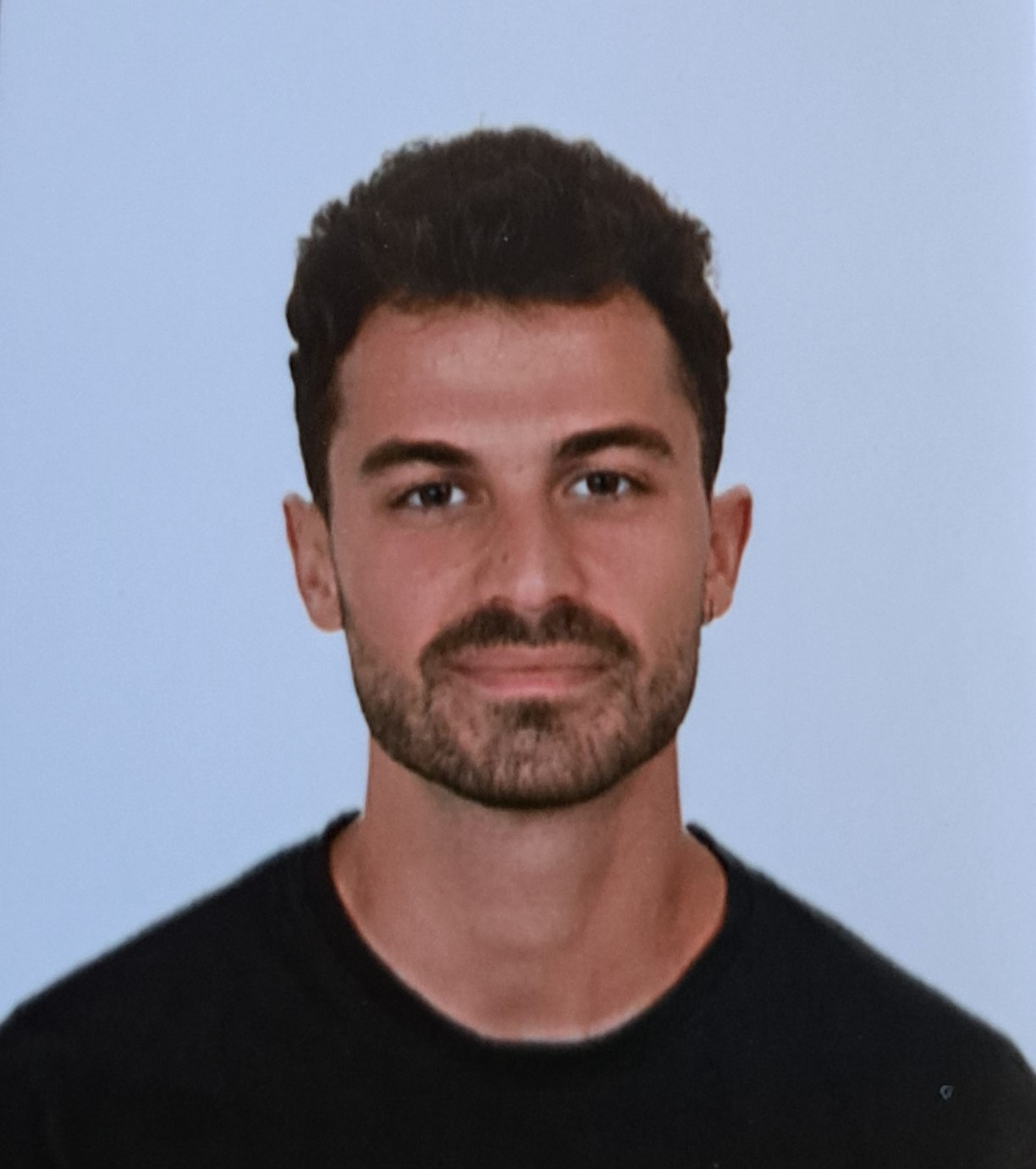}}]{Alberto Garz\'as-Villar} is a Ph.D. Candidate at the Delft University of Technology (The Netherlands). He is also affiliated with the Department of Socio-medical Sciences at Erasmus School of Health Policy \& Management, in Erasmus University Rotterdam (The Netherlands). He obtained his B.Sc. degree in mechanical engineering at UCLM (Ciudad Real, Spain) and his M.Sc. in biomedical engineering at UPV (Valencia, Spain). His research focuses on understanding the effect of patient diversity on the effectiveness of robotic motor learning.
\end{IEEEbiography}

\vspace{11pt}

\begin{IEEEbiography}[{\includegraphics[width=1in,height=1.25in,clip,keepaspectratio]{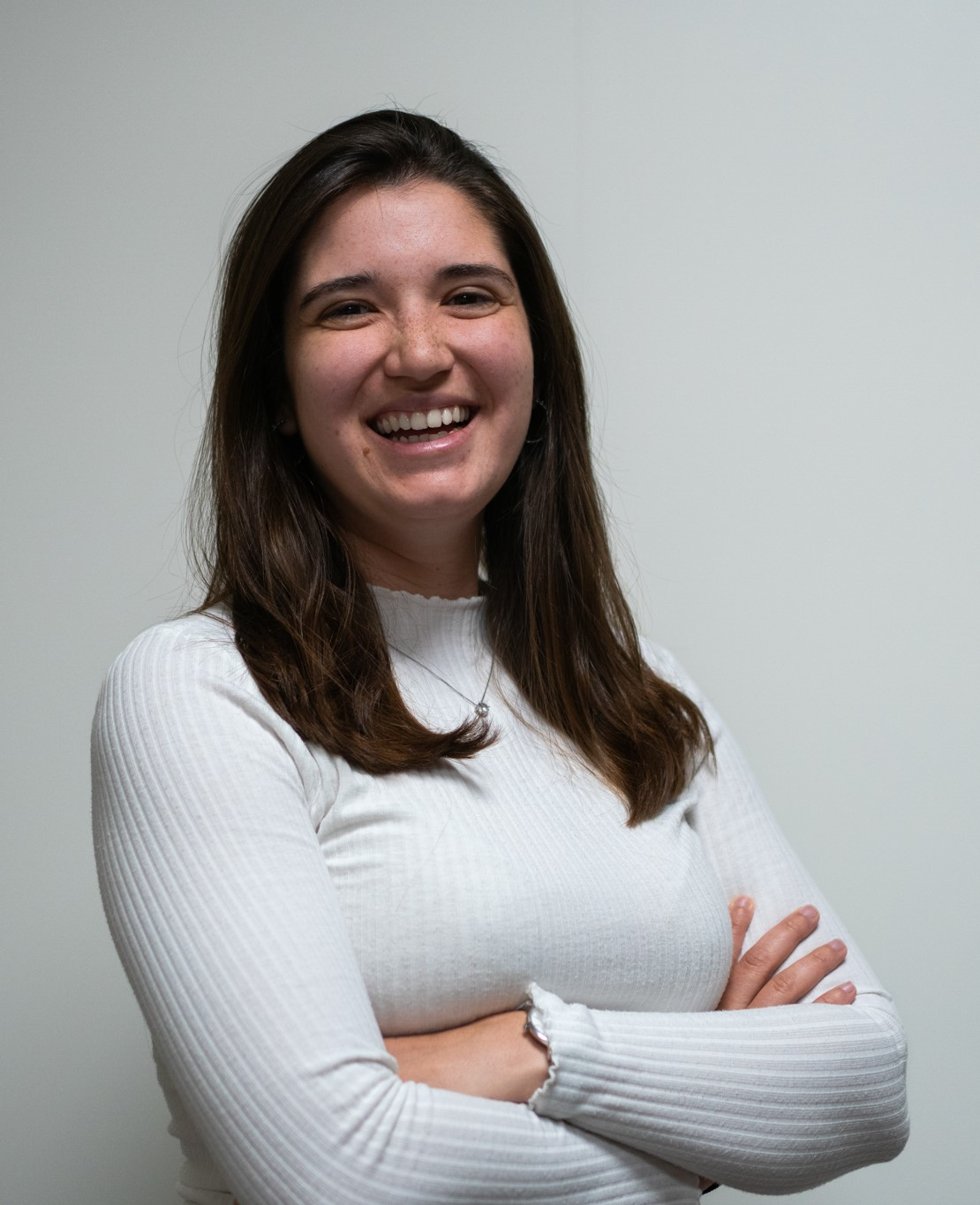}}]{Alba Riera-Cardona} is currently a Data Scientist at Zander Labs (The Netherlands), working in the field of passive Brain–Computer Interfaces. She obtained her B.Sc. degree in Biomedical Engineering from the University of Barcelona (Spain) in 2020 and worked as a software engineer at Creu Blanca Hospital from 2021-2022. She obtained her M.Sc. degree in Biomedical Engineering from the Delft University of Technology (The Netherlands) in 2025. Her master’s thesis was carried out at the Department of Cognitive Robotics and focused on leveraging robotics and haptic rendering for motor learning of force generation.
\end{IEEEbiography}

\vspace{11pt}

\begin{IEEEbiographynophoto}{Alexis Derumigny}
is an Assistant Professor in the Department of Applied Mathematics at TU Delft (The Netherlands). He received his Ph.D. in applied mathematics from the National School of Statistics and Economic Administration (ENSAE Paris) in 2019.
\end{IEEEbiographynophoto}

\vspace{11pt}

\begin{IEEEbiography}[{\includegraphics[width=1in,height=1.25in,clip,keepaspectratio]{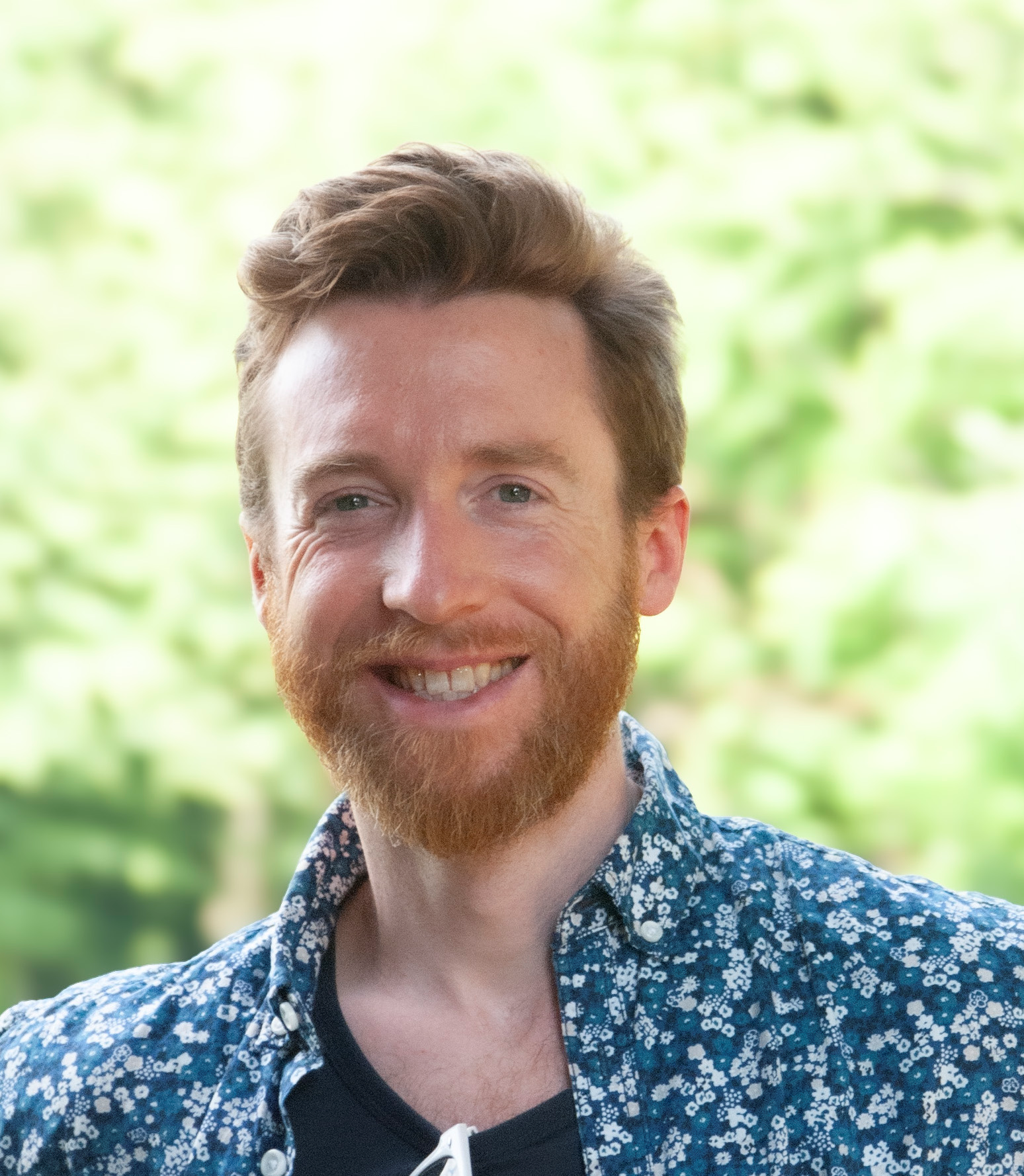}}]{J. Micah Prendergast} is an Assistant Professor at Delft University of Technology (The Netherlands). He obtained his M.Sc. and Ph.D. degrees from the University of Colorado at Boulder (USA). He was a postdoctoral researcher at Delft University of Technology from 2020-2023 prior to becoming an Assistant Professor in 2023. His research focus is on human-aware robotic systems for human-robot collaboration and assistance during rehabilitation, surgical interventions and physical work-processes.
\end{IEEEbiography}

\vspace{11pt}

\begin{IEEEbiography}[{\includegraphics[width=1in,height=1.25in,clip,keepaspectratio]{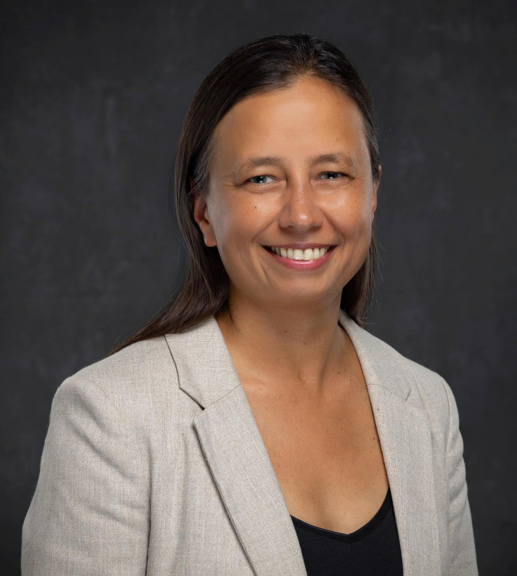}}]{Jane Murray Cramm} is Dean of the Tilburg School of Social and Behavioral Sciences at Tilburg University (The Netherlands) and Full Professor of Person-Centred Care at Tilburg University. She received the Ph.D. degree in Socio-Medical Sciences from Erasmus University Rotterdam. Her research focuses on person-centred care, diversity, and inclusive approaches to complex care challenges, particularly for vulnerable and ageing populations. She applies mixed-methods designs, including co-creation and participatory approaches, to develop and evaluate integrated, community-based care models, with attention to intersectionality and equity. Her work emphasizes engaged research, collaborating with citizens, professionals, and policymakers to ensure relevance, inclusivity, and societal impact. She is a Member of the Council for Quality at the National Health Care Institute and a member of the Supervisory Board of Mijzo.
\end{IEEEbiography}

\vspace{11pt}

\begin{IEEEbiography}[{\includegraphics[width=1in,height=1.25in,clip,keepaspectratio]{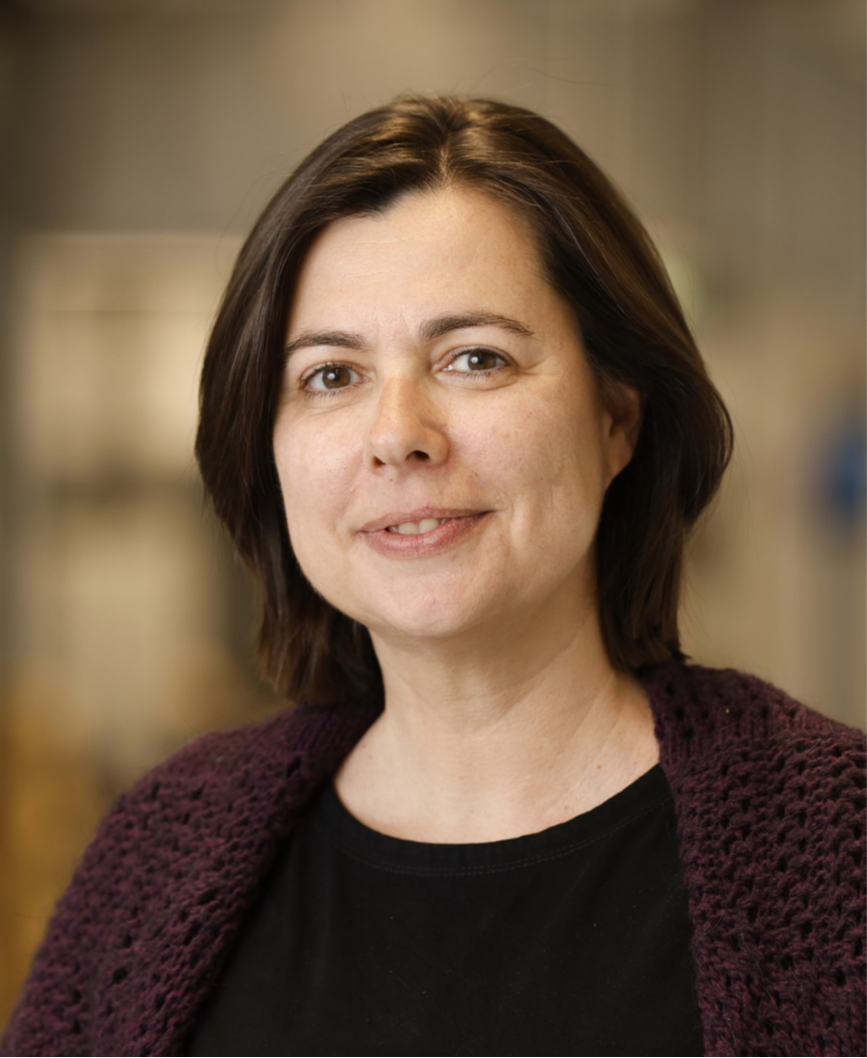}}]{Laura Marchal-Crespo} is an Associate Professor at the Delft University of Technology (The Netherlands) and Erasmus MC (The Netherlands). She obtained her M.Sc. and Ph.D. degrees from the University of California at Irvine (USA). She was a postdoc researcher at ETH Zurich and an Assistant Professor at the University of Bern (Switzerland). She became an Associate Professor at the Delft University of Technology in 2020, where she is leading the Human-Robot Interaction Group. She carries out research in the areas of human-machine interfaces and biological learning, and, specifically, in the use of robotics and virtual reality to aid people to rehabilitate after neurologic injuries.
\end{IEEEbiography}

\EOD

\end{document}